\documentclass[11pt]{article}

\usepackage[preprint]{acl}

\usepackage{times}
\usepackage{latexsym}

\usepackage[T1]{fontenc}

\usepackage[utf8]{inputenc}

\usepackage{microtype}

\usepackage{inconsolata}

\usepackage{graphicx}

\usepackage{hyperref}
\usepackage{url}
\usepackage{booktabs}
\usepackage{ulem}
\usepackage{listings}
\usepackage{svg}
\usepackage{multirow}
\usepackage{array}
\usepackage{threeparttable}
\usepackage[table]{xcolor}
\usepackage{tabularx}
\definecolor{openai}{RGB}{220,240,255}
\usepackage{makecell}
\usepackage{CJKutf8}
\usepackage{xcolor}
\usepackage{enumitem}
\usepackage{amsmath}
\setlist[itemize]{nosep, leftmargin=*}
\title{Meeseeks: A Feedback-Driven, Iterative Self-Correction Benchmark evaluating LLMs' Instruction Following Capability}

\author{
  Jiaming Wang$^{1}$, Yunke Zhao$^{1}$, Peng Ding, Jun Kuang$^{1}$, Yibin Shen$^{1}$, Zhe Tang$^{1}$, Yilin Jin, \\
\textbf{Zongyu Wang}$^{1}$, \textbf{Xiaoyu Li}$^{1}$,
\textbf{Xuezhi Cao}$^{1}$, \textbf{Xunliang Cai}$^{1}$ \\
  \\
  $^{1}$Meituan\\
  \texttt{\{wangjiaming15\}@meituan.com} \\
}

\begin{document}

\definecolor{openai}{RGB}{240,248,255}        
\definecolor{anthropic}{RGB}{255,245,238}      
\definecolor{deepseek}{RGB}{245,255,245}       
\definecolor{qwen}{RGB}{255,255,240}           
\definecolor{bytedance}{RGB}{245,245,255}      
\definecolor{google}{RGB}{255,251,240}         
\definecolor{moonshotai}{RGB}{251,240,255}     
\definecolor{zai}{RGB}{240,255,255}            

\maketitle
\begin{abstract}
The capability to precisely adhere to instructions is a cornerstone for Large Language Models (LLMs) to function as dependable agents in real-world scenarios. However, confronted with complex prompts, LLMs frequently encounter difficulties in fulfilling all specified requirements within a single response.
Drawing inspiration from recent advancements in Chain-of-Thought (CoT) prompting and self-correction methodologies, we introduce \textbf{Meeseeks}\footnote{The name is inspired by Mr. Meeseeks from ``Rick and Morty,'' a character renowned for efficiently accomplishing assigned tasks. See: \url{https://en.wikipedia.org/wiki/Mr._Meeseeks}},  a fully automated iterative instruction-following benchmark equipped with an integrated feedback mechanism. Meeseeks identifies erroneous components in model responses and provides corresponding feedback accurately, thereby iteratively guiding the model toward self-correction. 
The dataset contains over 700 curated instances annotated by 32 distinct capability tags in Chinese and English. 
Extensive experimental results reveal that different state-of-the-art commercial and open-source LLMs exhibit vastly disparate performance, and even after 20 turns of iterative feedback-driven self-correction, nearly all models demonstrate suboptimal performance. We conducted comprehensive analysis and uncovered numerous common issues prevalent in current state-of-the-art models, as well as several counterintuitive phenomena. Meeseeks has been open-sourced  on \url{https://anonymous.4open.science/r/Meeseeks-156A}.
\end{abstract}

\section{Introduction}
\label{sec:Introduction}
LLM agents have become essential tools in various applications, from customer service to content creation~\citep{intro1.1,intro1.2,intro1.3,intro1.4,intro1.5}. However, as their use expands, the instructions users provide are often complex and multifaceted, posing challenges for accurate execution~\citep{intro2.1,intro2.3,intro2.4,intro2.5,intro2.7,intro2.8}. Instruction following, or the ability to execute tasks based on natural language commands accurately, is critical for LLMs to be reliable in real-world scenarios. For instance, in healthcare, an LLM might need to generate reports with specific word counts or mention key terms multiple times. Misinterpreting such instructions could lead to errors with serious consequences~\citep{li2023chatdoctormedicalchatmodel,qiu2024llmbased}. Similarly, in finance, precise adherence to instructions ensures compliance and accuracy, avoiding potential risks~\citep{arun2023numericalreasoningfinancialreports,kim2025financialstatementanalysislarge}. These content creation tasks require LLMs to have strong text instruction-following capability. To evaluate the instruction-following ability of today's LLMs, researchers employ instruction-following benchmarks to determine how effectively LLMs can align with human intentions. This research indicates that models are unable to address problems flawlessly in scenarios where prompts contain numerous constraints.

Chain-of-Thought (CoT) prompting has been primarily developed to enhance the reasoning capabilities of LLMs in mathematical and logical tasks by guiding them to generate intermediate steps~\citep{wei2022chain,wang2022towards,yu2023towards,zhang2022automatic}. Building upon CoT, self-correction techniques enable LLMs to identify and rectify errors in their reasoning processes. ~\citep{zhuang2024training,huang2023large,liu2024large,schiefer2023capacity,mao2024intrinsic}. 
While existing self-correction methods are primarily evaluated on mathematical or logical problems, we extend this approach to multi-constraint instruction-following tasks to probe the upper limits of LLMs' instruction-following capabilities.

Therefore, we introduce Meeseeks, a multi-turn automatic instruction-following benchmark with feedback system. The framework enables Meeseeks to fully automatically identify specific constraints that the LLM fails to meet and explicitly explain the reasons for such failures. The system operates in successive rounds, where Meeseeks evaluates the LLM response, provides feedback, and then assesses the resulting self-correction. Moreover, Meeseeks applies a more challenging dataset embedding 32 capability tags and 3 cognitive evaluation dimensions (intent recognition, granular content validation and output structure validation), examining the entire thought process of the model to establish an integrated instruction-following ability evaluation system. Furthermore, due to the high cost of single to multiple turns evaluation and more challenging data, Meeseeks introduces a code-guided, rule-augmented LLM-based evaluation to reduce cost and raise accuracy of the evaluation workflow. 
Meeseeks reveals nuanced patterns in LLMs' multi-turn self-correction abilities that challenge conventional understanding of instruction following as well as a significant gap between different LLMs' single-attempt instruction-following performance and their upper-limit instruction-following capability. Even after 20 rounds of self-correction, all prevailing LLMs still fail to provide responses that satisfy all constraints. In conclusion, our main contributions are:

\begin{itemize} \item We introduce Meeseeks, a challenging multi-turn automatic instruction-following benchmark that operates in successive evaluation-feedback-correction cycles across 32 capability tags and 3 evaluation dimensions. The framework accurately identifies constraint violations, provides targeted feedback, and evaluates LLMs' self-correction attempts.
\item We present an enhanced automatic evaluation pipeline by introducing a novel code-guided, rule-augmented LLM-based evaluation framework. Experimental results demonstrate that our proposed method significantly outperforms existing approaches, delivering substantial improvements in evaluation accuracy, raising accuracy from 78.7\% to 98.4\%.
\item We conduct experiments across 17 state-of-the-art models and observe that despite implementing 20 iterations of feedback-driven self-correction mechanisms, both leading commercial and open-source LLMs consistently failed to achieve accuracy rates exceeding 91\%. Through systematic analysis, we identify both common and distinct limitations among these models, thereby contributing valuable insights and establishing reference benchmarks for future model development and optimization efforts.

\end{itemize}

\section{Related Works}
\label{sec:related_works}

IF-Eval~\citep{single_benchmark_IFEVAL} pioneered comprehensive instruction-following evaluation by introducing rule-verifiable instructions (e.g., ``Include \texttt{keyword1}, \texttt{keyword2} in your response''). While this approach enables automatic verification, the dataset's simplicity and synthetic nature limit its applicability to real-world scenarios. Subsequent benchmarks, such as InFoBench~\citep{single_benchmark:InFoBench}, CELLO~\citep{single_benchmark_CELLO}, FollowBench~\citep{multi_benchmark:followbench}, and COLLIE~\citep{single_benchmark_COLLIE}, advanced the field by employing constraint-based frameworks to handle more complex data. However, these methods still rely on specific format constraints for evaluation.
Complexbench~\citep{single_benchmark_Complexbench} effectively addresses these limitations by leveraging rule-augmented, LLM-based evaluation processes. This approach eliminates the need for extensive prompt restrictions, allowing for more flexible response formats from the evaluated LLMs. Current instruction-following benchmarks are predominantly designed for single-turn interactions and fail to capture the complexity of multi-turn conversational scenarios, thereby limiting their effectiveness in comprehensively evaluating models' instruction-following capabilities. Parrot~\citep{multi_benchmark_Parrot} introduced a framework for collecting human-like multi-turn instructions that mimic natural dialogue patterns. Multi-IF~\citep{multi_benchmark_Multi-IF} extended IF-Eval to multi-turn sequences across multiple languages, revealing performance degradation in later turns and in non-Latin scripts. StructFlowBench~\citep{multi_benchmark_StructFlowBench} proposed a structured framework with six types of inter-turn relationships to evaluate dialogue coherence and contextual understanding. However, these multi-turn benchmarks typically allow LLMs only a single attempt at each turn, introducing new requirements in subsequent turns. In contrast, Meeseeks allows models self-correct their response as the turn iterates.

\section{Meeseeks}
\begin{figure*}[t]
  \includegraphics[width=1\linewidth]{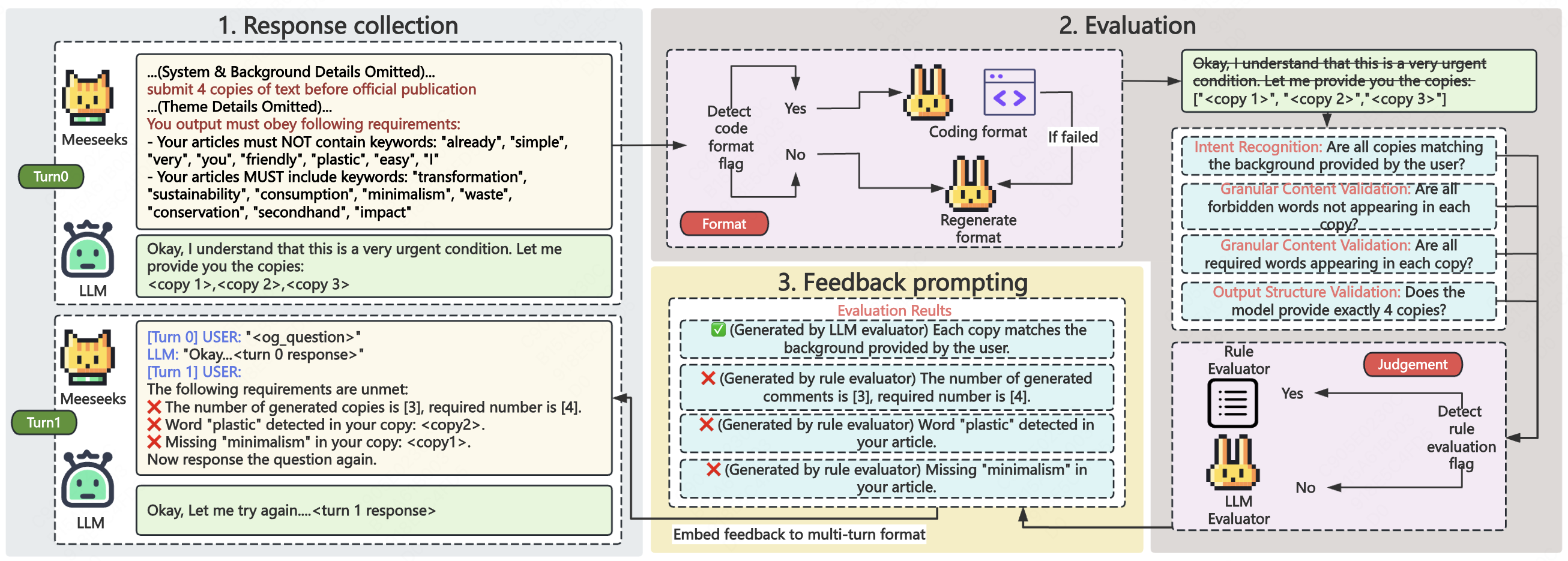} \hfill
    \caption{One complete Meeseeks iteration includes response collection, evaluation, and feedback prompting.}
    \label{fig:fig1}
\end{figure*}

Meeseeks implements an iterative self-correction framework comprising response collection, evaluation, and feedback prompting (Figure~\ref{fig:fig1}). More specifically, given questions with multiple constraints, each iteration follows three steps: (1) collecting responses from the model while enforcing all constraints, (2) evaluating the response against each constraint, and (3) providing constraint-specific evaluation results and explanations as feedback to prompt the model for self-correction in the subsequent iteration.
The following sections provide a comprehensive overview of Meeseeks. Section~\ref{sec:eval_pip} and Section~\ref{sec:feedback_prompting} provide details of evaluation and feedback prompting respectively; Section~\ref{sec:LLM-format} explains our code-guided, rule-augmented, LLM-based evaluation addressing cost and accuracy challenges in multi-turn frameworks, serving the evaluation; 
Section~\ref{sec:dataset} presents the quality control and the structure of our dataset;
Section~\ref{sec:metrics} elaborates on metrics used in our experiments.

\subsection{Evaluation}
\label{sec:eval_pip}
Due to hallucinations in LLM, directly relying on them to determine whether each constraint is satisfied by the tested model's response is unreliable~\citep{Huang2024,li2023halueval,zhu2024haluevalwild,gu2024medvh,qiu2024longhalqa,chen2024diahalu,Hao2025,bang2025hallulens,yan2024evaluating}.
Therefore, our primary method for judging constraints consists of two steps: first, use an LLM evaluator to format the model’s response (typically by converting it into a Python list); second, employ a Python script to assess whether each constraint is correctly satisfied. To facilitate the format section, the framework integrates a dual format mechanism: the coding format, which activates the code-specific pipeline when coding flags are present (flag detail:  Appendix~\ref{sec:data_structure}; coding format detail: Section~\ref{sec:LLM-format}), and the regenerate format, which formats model responses through regeneration. 
The evaluation procedure is bifurcated according to the rule labels associated with each constraint. For constraints with null rule labels (see Appendix~\ref{sec:data_structure}), the LLM evaluator assesses the response directly, without any formatting. When specific rule labels are present, corresponding rule-based Python judgment scripts are invoked to verify the validity of the formatted content.

\subsection{Feedback Prompting}
\label{sec:feedback_prompting}
After evaluation, each constraint is accompanied by a judgment explanation; if a constraint is unmet, the feedback specifies the reason for failure. With the exception of a small portion of feedback generated directly by the LLM evaluator, the majority of feedback is automatically produced by the constraint-specific scripts. These evaluation results and corresponding feedback are integrated into the tested LLM's chat template for subsequent turns, serving as targeted guidance for its self-correction. The framework monitors improvement after each self-correction iteration.

\subsection{Code-guided, Rule-augmented LLM-based Evaluation}
\label{sec:LLM-format}
Addressing the limitations of rule-verifiable instructions and format constraints problem mentioned in Section \ref{sec:related_works}, the prevailing method is to employ an LLM-based approach to eliminate redundant prompt constraints. This approach identifies and retains core content by eliminating auxiliary text, followed by rule-based evaluation to ensure compliance with specific constraints. Consider the Prompt: \textit{Write me a research report about LLM, make sure the report should include less than 2000 words,} and the example problem shown in Table~\ref{tab:model_response_format}.

\begin{table}[h]
\tiny
\renewcommand{\arraystretch}{0.9}
\centering
\begin{tabular}{p{7.5cm}}
\toprule
\textbf{$\circ$ Raw Model Response} \\
\midrule
\textbf{Response:} Okay, I will write you a 2000-word essay. LLM is...(1998 words). I hope you like the answer. \\
\textbf{Check:} Word count: 2014 > 2000 $\rightarrow$ Constraints unmet \textbf{[INCORRECT]} \\
\addlinespace
\toprule
\textbf{$\circ$ Formatted Model Response} \\
\midrule
\textbf{Response:} \sout{Okay, I will write you a 2000-word essay.} LLM is...(1998 words) \sout{I hope you like the answer.} \\
\textbf{Check:} Word count: 1998 < 2000 $\rightarrow$ Constraints met \textbf{[CORRECT]} \\
\bottomrule
\end{tabular}
\caption{Raw model responses can introduce evaluation errors during Python script execution. The prevailing method regenerates the model response and cleans the redundant texts.}
\label{tab:model_response_format}
\end{table}

However, this approach encounters significant challenges in both computational efficiency and accuracy. For long model responses (such as extended essays) and large-scale multi-element generation, the LLM-based format extraction requires regenerating entire clean texts to remove non-essential components, resulting in substantial computational overhead. Furthermore, format accuracy degrades considerably as the complexity of model responses increases, creating a scalability bottleneck for practical applications. To address these issues, we propose a code-guided, rule-augmented LLM-based evaluation. Specifically, we use prompts to guide the LLM to format model response by coding, rather than having the model regenerate all context. By applying the function created by the LLM Evaluator, the necessary information can be formatted efficiently without the regeneration step 
Table~\hyperref[tab:cgrbeval]{2}). Through experimental validation, we demonstrate that our code-guided formatting delivers substantial improvements in evaluation accuracy with performance gains reaching 97.46\%, while achieving up to 71.08\% reduction in token consumption compared to prevailing formatting approaches (See Section~\ref{sec:cgrexperiment}).

\subsection{Data collection, Taxonomy and Quality}
\label{sec:dataset}
Meeseeks comprises over 700 meticulously curated instances in Chinese and English, annotated with 32 distinct capability tags, and with a substantially greater difficulty compared to existing datasets in the domain. The dataset is collected from our business log and rigorously formatted to templates (Appendix~\ref{sec:datatemplate}), wherein all constraints are manually annotated by professional annotators to ensure both precision and contextual relevance. Similar to prevailing instruction following benchmarks, the prompts from our log also contain multiple output constraints, and we use a cognitive-inspired framework that decomposes instruction-following into three fundamental dimensions (Figure~\ref{fig:fig1}): Intent Recognition, Granular Content Validation, and Output Structure Validation. This framework employs hierarchical capability tags across 3 levels and 32 capability tags in total, with lower levels subordinate to higher ones (Appendix~\ref{sec:appendixB}). Data instances are systematically generated by varying template parameters and synchronously updating the associated constraint variables, with \textbf{NO involvement of LLM at any stage of data generation}. To further enhance the reliability and objectivity of the evaluation, all subjective capability constraints—such as those assessing emotional intelligence in model responses—are deliberately excluded from the dataset. For quality assurance, we recruit three experienced annotators, each holding at least a bachelor's degree with professional expertise in data annotation, and conducted comprehensive reviews of every data instance and its associated constraints, rigorously verifying their validity and consistency. For instance, they identified and resolved logical contradictions, such as cases where one constraint required the inclusion of an "idea" while another simultaneously prohibited a "great idea," thereby ensuring the integrity and coherence of the synthesized data.

\begin{table}[htbp]
\tiny
\renewcommand{\arraystretch}{0.9}
\centering
\begin{tabular}{p{7.5cm}}
\toprule
\textbf{$\circ$ LLM Response} \\
\midrule
Due to the length of the report, the following is a simplified version of the research report outline, which you can further expand as needed:---\# Research Report: \textbf{System Spam Problems and Solutions Faced by Ordinary Users\#\# Summary This report collects data for ordinary users in..... (2000 words omitted).} ---Please note that this is only a simplified version of the report outline. \\
\addlinespace
\toprule
\textbf{$\circ$ Prevailing Formatting} \\
\midrule
\textbf{System Spam Problems and Solutions Faced by Ordinary Users\#\# Summary This report collects data for ordinary users in..... (2000 words omitted).} \\
\texttt{2,754 Tokens} \\
\addlinespace
\toprule
\textbf{$\circ$ Code-guided Formatting} \\
\midrule
\vspace{-0.4cm}
\begin{lstlisting}[
    language=Python,
    basicstyle=\tiny\ttfamily\color{black},  
    keywordstyle=\color{blue!70!black},
    commentstyle=\color{green!45!black},
    stringstyle=\color{red!55!black},
    showstringspaces=false,
    breaklines=true,
    breakatwhitespace=true,
    frame=single,
    framesep=1pt,
    basewidth=0.45em,
    columns=flexible,
    backgroundcolor=\color{gray!3}]
def extract_info_list(model_response):
    """
    Because the object to be captured is the entire
    research report outline. Therefore: After capturing
    "[# Research report: ]", the content between "
    [.---Please note that]" is enough.
    """
    cleaned_text = re.sub(
        r'^.*# Research Report: |.---Please note that.*$', '', 
        model_response, 
        flags=re.DOTALL)
    return [cleaned_text]
\end{lstlisting}
\texttt{98 Tokens} \\
\bottomrule
\end{tabular}
\caption{Our code-guided, rule-based evaluation instructs LLMs to generate code for formatting model response, thereby replacing the regeneration techniques employed in existing methodologies.}
\label{tab:cgrbeval}
\end{table}

We conduct a comprehensive evaluation accuracy on Meeseeks' own dataset. We have our annotators independently assess responses generated by Claude-3.7-Sonnet, GPT-4o-20241120, and DeepSeek-R1. Given the straightforward nature of the evaluation tasks for human assessors, we are confident in the annotators' ability to provide accurate evaluations of the LLM responses.
Following the collection of evaluation results, we analyze inter-annotator agreement and observed over 99\% consensus among the three annotators, demonstrating strong reliability in our human evaluation framework. Any remaining discrepancies were manually reviewed to ensure final quality assurance. Through this rigorous evaluation process, we achieve a substantial improvement in end-to-end accuracy from 78.7\% (The prevailing format accuracy on the Meeseeks dataset) to 98.4\%.

\subsection{Metrics}
\label{sec:metrics}
Serving as trustworthy agents, LLMs must fulfill all requirements specified in the user's prompt for the responses to be considered usable. \textbf{Utility rate} reflects the ratio of usable responses to total evaluated responses and is essential for evaluating an LLM's effectiveness as a practical and professional agent. Additionally, to help researchers identify subtle differences in instruction-following abilities between different LLMs, we propose the Meeseeks score, which indicates the overall capability tag accuracy of the LLM (see formula in Appendix~\ref{sec:metrics_formula}). Furthermore, Meeseeks generates a capability report with accuracy statistics across various capability levels, enabling researchers to understand the instruction-following proficiency of LLMs from different perspectives (sample report in Appendix~\ref{sec:sample_capa_report}). The accuracy metric denotes the utility rate hereafter.

\section{Experiments}
\label{sec:expers}
\subsection{Experiments on Code-guided, Rule-augmented LLM-based Evaluation}
\label{sec:cgrexperiment}
\begin{figure}[!htb]
    \includegraphics[width=0.5\textwidth]{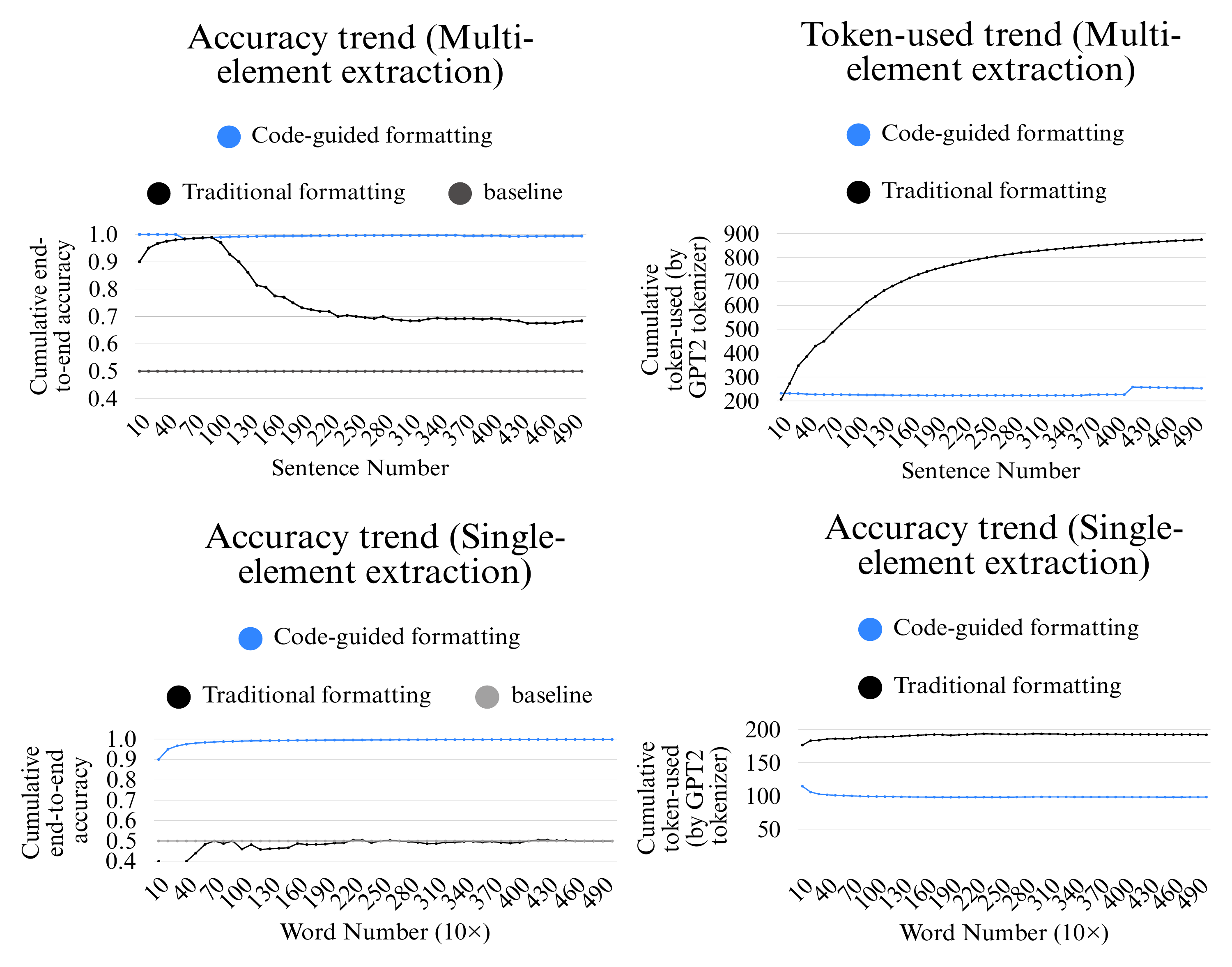}
    \caption{Our code-guided formatting demonstrates remarkably consistent performance in both token utilization and accuracy metrics.}
    \label{fig:fig2}
\end{figure}

To evaluate the effectiveness of our code-guided, rule-based evaluation, we conduct experiments in two distinct scenarios: long single element generation, where the model is tasked with producing a complete article, and large-scale multi-element response generation, where the model is required to generate a substantial number of comments (These two frequently used scenarios are utilized and modified from Complexbench's dataset). For long single element scenarios, we choose the question: \textit{A man is performing on stage for the audience, ... Please continue and expand this passage with at least 20 sentences.} We vary the required number of sentences from 0 to 500 with an increment of 1, generating synthetic data.
For large-scale multi-element scenarios, we choose the question: \textit{People are ... Please write an essay. Requirements: Argumentative essay, objective and neutral language style, no less than 800 characters.} Similarly, we set the required character count from 50 to 5000 (in increments of 100, and rather than use "less than", output length is required in a certain range), generating synthetic data and answers.
The ratio of correct to incorrect answers in both scenarios is set to 1:1, and employed Qwen2.5-32B + Qwen2.5-32B-Coder ~\citep{EVAL_LLM_qwen2_5} as the evaluator (also for main experiments). Subsequently, we conduct a comparative analysis of the accuracy and token consumption patterns exhibited by the code-guided and prevailing formatting methods (Figure~\ref{fig:fig2}).

In large-scale multi-element scenarios, the code-guided formatting approach exhibited substantial performance gains over the prevailing method. It attained an average end-to-end accuracy of 0.994 with an average token usage of 252.94, compared to the prevailing formatting's accuracy of 0.684 and token usage of 874.65—yielding a 45.32\% improvement in accuracy and a 71.08\% reduction in token consumption. In scenarios with long single element, code-guided formatting sustained high efficacy, achieving an end-to-end accuracy of 0.998 and average token usage of 98.73, whereas the prevailing approach encountered severe hallucinations and often misidentified core content, resulting in an accuracy drop to the baseline level of 0.5 and token usage of 191.99; this translated to a 97.46\% accuracy enhancement for code-guided formatting. The limitations of prevailing formatting primarily arose from its token-saving strategy, which defaulted to outputting merely the "thinking process" and "ALL" without supplementary context ("ALL" indicates that the evaluator considers the model's response to be free of redundancy). Efforts to address this by employing advanced evaluators such as GPT-4o yielded no discernible benefits, suggesting that the underlying issue resides in the few-shot prompt architecture of the prevailing method. These outcomes highlight the enhanced robustness and resource efficiency of code-guided formatting in varied and demanding contexts.

\label{sec:evaluation}
\subsection{Main Experiments}
We evaluate 17 representative LLMs over 20 turns and find models from commercial APIs consistently demonstrate superior performance in both single-turn and multi-turn dialogue settings (When the input length surpasses the maximum context window capacity of the evaluated LLM, the instance is automatically classified as failed). Notably, o3-mini (high) achieves the highest performance from turn 1 through turn 5. In subsequent turns, Claude Sonnet 4 Thinking surpasses other models, establishing itself as the best-performing system in extended interactions. However, despite 20 turns of feedback-guided self-correction, the performance of most models remained suboptimal. In the following sections, we provide a systematic interpretation of the phenomena presented in Table~\hyperref[tab:MAINTABLE]{3}. Further analysis is provided in following sections. \citep{LLM_claude_sonnet4, LLM_qwen3, LLM_doubao2024, LLM_dsv3, LLM_gemini_flash_2024, LLM_glm, LLM_gpt4_1, LLM_kimik2, LLM_o3mini}

\begin{threeparttable}
\scriptsize
\begin{tabular}{>{\columncolor{white}}l r r r r}
\toprule
\textbf{Model ↓ \quad \quad Turn →} & \makecell{\textbf{1-5}\\\textbf{avg}} & \makecell{\textbf{6-10}\\\textbf{avg}} & \makecell{\textbf{11-15}\\\textbf{avg}} & \makecell{\textbf{16-20}\\\textbf{avg}} \\
\midrule
\midrule
\multicolumn{1}{>{\columncolor{deepseek}}l}{DeepSeek-V3.1*} & .567 & .745 & .783 & .789 \\
\multicolumn{1}{>{\columncolor{zai}}l}{zai-org/GLM-4.5*} & .637 & .805 & .84 & .854 \\
\multicolumn{1}{>{\columncolor{bytedance}}l}{Doubao Seed 1.6} & .531 & .695 & .744 & .766 \\
\multicolumn{1}{>{\columncolor{moonshotai}}l}{Kimi-K2-Instruct*} & .510 & .595 & .619 & .623 \\
\multicolumn{1}{>{\columncolor{google}}l}{google/Gemini 2.5 Flash} & .522 & .636 & .668 & .671 \\
\multicolumn{1}{>{\columncolor{openai}}l}{openai/GPT 4.1} & .550 & .634 & .673 & .689 \\
\multicolumn{1}{>{\columncolor{openai}}l}{o3-mini (high)} & .715$^\dagger$ & .813 & .831 & .842 \\
\multicolumn{1}{>{\columncolor{anthropic}}l}{Claude Sonnet 4} & .584 & .747 & .783 & .791 \\
\multicolumn{1}{>{\columncolor{anthropic}}l}{Claude Sonnet 4 Thinking} & .656 & .832$^\dagger$ & .88$^\dagger$ & .899$^\dagger$ \\
\multicolumn{1}{>{\columncolor{anthropic}}l}{Claude Opus 4} & .622 & .787 & .822 & .837 \\
\multicolumn{1}{>{\columncolor{anthropic}}l}{Claude Opus 4 Thinking} & .670 & .814 & .858 & .871 \\
\multicolumn{1}{>{\columncolor{qwen}}l}{Qwen3-235B-A22B*} & .567 & .697 & .727 & .734 \\
\multicolumn{1}{>{\columncolor{qwen}}l}{Qwen3-235B-A22B-think*} & .663 & .826 & .852 & .866 \\
\multicolumn{1}{>{\columncolor{qwen}}l}{Qwen3-32B*} & .479 & .55 & .565 & .57 \\
\multicolumn{1}{>{\columncolor{qwen}}l}{Qwen3-32B-think*} & .500 & .617 & .642 & .657 \\
\multicolumn{1}{>{\columncolor{qwen}}l}{Qwen3-14B*} & .418 & .462 & .47 & .47 \\
\multicolumn{1}{>{\columncolor{qwen}}l}{Qwen3-14B-think*} & .466 & .549 & .567 & .573 \\
\bottomrule
\end{tabular}
\label{tab:MAINTABLE}
\begin{tablenotes}
\footnotesize
\item[*] All values omit leading zero (average on 3 runs). Models marked with * are open-source.
\item[$\dagger$] Indicates the highest score in each turn.
\end{tablenotes}
\caption{Utility rate over 20 turns (Complete result in Appendix~\ref{sec:complete_experiments_results})}
\end{threeparttable}

\section{Analysis}
\subsection{Performance Gap On First and Last Turn}
\begin{figure}[!htb]
    \includegraphics[width=0.5\textwidth]{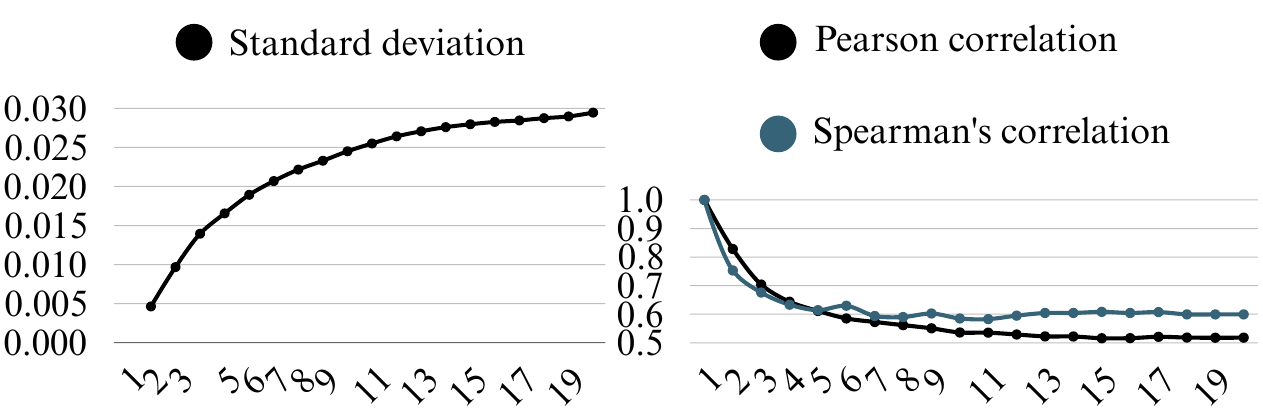}
    \caption{As the turns iterate, the standard deviation of utility rates gradually increases; The correlation between subsequent round rankings and turn 1 rankings shows a declining trend.}
    \label{fig:fig3}
\end{figure}

We design the experiment such that models were only required to correct questions they had previously answered incorrectly in subsequent turns. Theoretically, as the pool of questions that can be answered correctly diminishes, the performance gap between different models should gradually narrow. However, our findings indicate the opposite: the disparity between models actually widens with an increasing number of turns. Through this iterative process of correction over multiple turns, we can progressively approach the upper limit of a model's instruction-following capability. This phenomenon suggests that the difference in the upper limits of instruction-following abilities among various models is substantially greater than the single-turn performance differences might suggest (Figure~\ref{fig:fig3}). Concurrently, we observe that the rankings of the models evolve across turns, and their correlation with the initial ranking in the first turn progressively weakens. We calculate the Pearson and Spearman correlation coefficients between the model rankings in Turn 1 and those in each subsequent turn. The results reveal that the correlation between a model's performance in a single attempt and its upper limit of instruction-following ability initially diminishes rapidly and subsequently plateaus as the number of turns increases (Figure~\ref{fig:fig3}). This further reinforces our earlier conclusion: the disparity between a model's single-turn instruction-following performance and its ultimate capability ceiling is more significant than initially anticipated.

\vspace{-0.5cm}

\subsection{Performance Gap Across Capability Tags}
To further elucidate the underlying reasons for suboptimal model performance, we examine the identified issues through the lens of constraint satisfaction. Specifically, we analyze these challenges from the perspective of our three top-level capability tags along with some additional findings.

\begin{figure*}[t]
    \centering
    \includegraphics[width=1\textwidth]{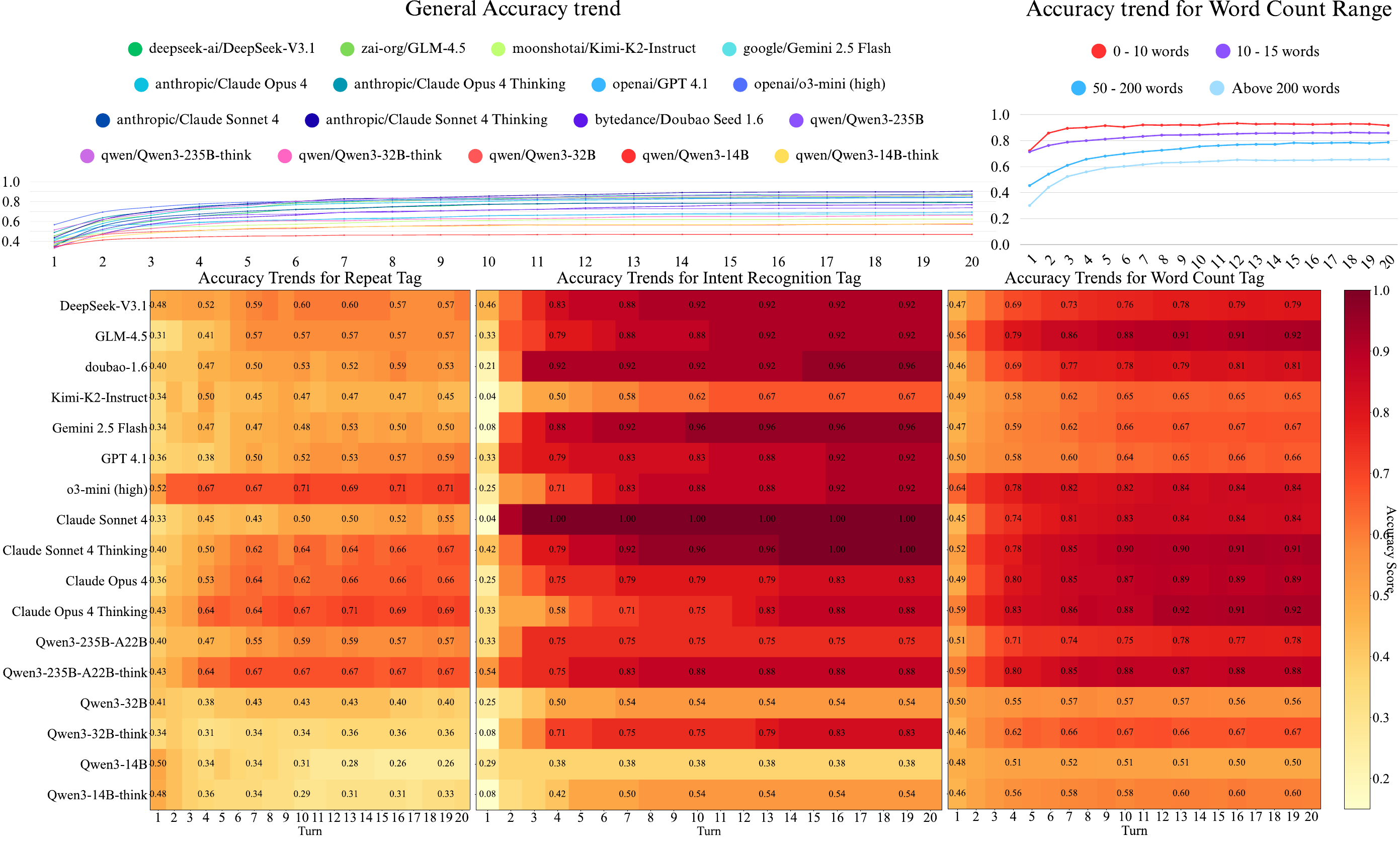}
    \caption{Performance trends varied significantly across models and capability tags.}
    \label{fig:allcapatrends}
\end{figure*}

\textbf{Output Structure Validation}\quad
When models are tasked with generating multiple contexts (e.g., producing numerous titles or names), their outputs exhibit substantial redundancy, with minimal improvement following feedback interventions. Despite explicit feedback indicating which content elements are duplicated, virtually all evaluated models demonstrate limited capacity to rectify such extensive redundancy issues (Figure~\ref{fig:allcapatrends}). In more demanding scenarios, when models are required to generate content adhering to specific proportional constraints—for instance, producing 20 pieces of copy with a long-form to short-form ratio of 3:2—we observe that most models fail to accurately determine the requisite quantities of long-form versus short-form content. However, reasoning contents in reasoning models help predetermine the quantities of long and short content within their reasoning processes, thereby mitigating such failures.

\textbf{Granular Content Validation}\quad
Within this dimension, model deficiencies are predominantly concentrated on word count accuracy. We observe that models fundamentally lack precise word counting capabilities; however, they should theoretically possess a general awareness of output length \citep{word_count1, word_count2, word_count3, word_count4}. Indeed, our experiment shows that models would adjust their output length to align with specified requirements through our feedback mechanism (Figure~\ref{fig:allcapatrends}). Nevertheless, we identify a critical phenomenon: when models receive feedback indicating that their output word count fails to meet requirements—accompanied by the actual word count—they frequently exhibit severe overcorrection behaviors. For instance, when the requirement specifies approximately 400 words (with acceptance criteria of 360-440 words), and the model's initial output contains 300 words, upon receiving feedback regarding the word count deficiency, most models initially make modest upward adjustments (e.g., to 330 words). However, upon receiving subsequent feedback indicating continued insufficiency, catastrophic overcorrection occurs: outputs may abruptly escalate to 600 words. This adjustment momentum exhibits bidirectional persistence, with subsequent iterations potentially plummeting to 100 words. While we observe that certain models can gradually resolve some questions through sustained iterative feedback, a substantial proportion of models remain unable to address this fundamental issue even after 20 rounds of feedback-guided iterations. To provide more granular analysis, we introduce detailed capability tags spanning: 0-10 words (precise), 10-50 words (precise/range), 50-200 words (range), and above 200 words (range) (Figure~\ref{fig:allcapatrends}). These results demonstrate that models exhibit markedly superior adherence to word count constraints at smaller magnitudes.

\textbf{Intent Recognition}\quad
In this dimension, we implement a particularly challenging task requiring models to modify prompts. Specifically, we delimit prompts using different special notation and instruct models to modify the enclosed prompt content either at the beginning or end of the interaction. We observe that the majority of models fail to comprehend our requirement for prompt modification on the initial attempt. Instead, they erroneously execute the tasks specified within the bracketed prompt content. Even reasoning-capable models demonstrate this misinterpretation during their initial cognitive processes, despite their enhanced reasoning capabilities. Subsequently, we employ feedback mechanisms to explicitly reveal our actual intention and emphasize the prompt modification requirement. However, we find that a substantial proportion of models remain unable to successfully modify the designated prompt content even after clarification. We attribute this limitation to fundamental optimization challenges inherent in multi-turn dialogue systems, suggesting that current model architectures may possess insufficient capacity for maintaining and adapting to task clarification across conversational iterations.

\textbf{Additional Findings}\quad
For regional knowledge requirements, we observe that broad constraints such as rhyming patterns and traditional-simplified Chinese character restrictions pose relatively manageable challenges for models. However, when confronted with more specialized linguistic requirements—such as generating content adhering to specific tonal pattern (pingze) constraints—models encounter substantial difficulties. Given that virtually all evaluated models lack a comprehensive understanding of tonal pattern principles, they demonstrate near-universal failure in completing pingze-related generation tasks. Notably, we identify a particularly intriguing phenomenon: Our feedback mechanism provides models with pingze information corresponding to their generated content. Among all evaluated models, o3-mini (high) emerges as the sole exception capable of leveraging the pingze information provided in feedback to reconstruct content that satisfies the specified tonal pattern requirements. This capability represents a remarkably encouraging discovery, demonstrating the big in-context learning gap among models.

\subsection{Reasoning/Non-reasoning Models Gap}
Either in single-turn or multi-turn scenarios, most reasoning models generally outperform their non-reasoning counterparts. These findings challenge the prevailing consensus that thinking models are less effective in instruction-following tasks~\citep{wang2025explicit}. Our analysis reveals that reasoning models demonstrate consistent improvements over non-reasoning models across five capability tags in the final evaluation round (Generate with certain number of word X; Generate in 50-200 words; Generate in 10-50 words; Generate in X times word number of reference text; Generate Chinese-English-mixed article).
While individual reasoning models exhibit minor performance degradations on specific metrics, these decreases are neither substantial nor systematic across models. Notably, most reasoning models show significant performance gains through feedback-driven self-correction that substantially exceed those of non-reasoning models, even when the reasoning models already demonstrate initial advantages in single-turn evaluations. More specifically, we find reasoning models demonstrate superior sensitivity to word number, whether for Chinese characters or English words. They exhibit greater robustness in both range-based and precise numerical outputs, effectively mitigating the catastrophic adjustment momentum issues previously identified. Moreover, in reasoning contents, reasoning models proactively calculate required content length and proportions for tasks requiring inferential steps, leading to improved task execution. For "Generate with certain number of word X" tasks, reasoning models consistently verify word occurrence within their reasoning context, eliminating the "random guessing" behaviors observed in non-reasoning models.

\section{Conclusion}
We introduce Meeseeks, a multi-turn instruction-following benchmark that evaluates LLMs' capability to satisfy multi-constraint instructions through iterative feedback. Facing accuracy drop in complex scenarios, we propose code-guided rule-augmented LLM-based evaluation framework, raising the evaluation accuracy from 78.7\% to 98.4\%. Through our experiments across 17 popular models, we find even after 20 turns of self-correction, almost all LLMs fail to exceed 90\% accuracy, highlighting fundamental limitations in LLMs' instruction-following capabilities. The substantial performance gaps between single-turn capabilities and multi-turn upper bounds reveal that different LLMs exhibit varying instruction-following ceilings. Consequently, we provide rigorous analysis of the self-correction mechanisms and offer new insights for developing more reliable instruction-following systems in future LLM research.

\section*{Limitations}
In the comparison between reasoning and non-reasoning models, while adding "Think step by step" prompting to non-reasoning models might potentially improve their performance scores, we deliberately refrained from employing this approach. We did not explicitly restrict models from outputting auxiliary reasoning content, and our rationale for this decision stems from the benchmark's fundamental purpose of reflecting authentic model capabilities. Considering that users typically lack professional prompting expertise and would experience suboptimal user experience when extracting answers from extensive reasoning contexts, reasoning models prove more suitable for end-users under our experimental conclusions. Consequently, we did not conduct such granular ablation studies.

Instruction-following evaluation is inherently constrained by its design to disregard answer correctness. Even with increased task complexity, establishing a direct correlation between model performance and user experience remains challenging. In our future work, we aspire to bridge the gap between instruction-following capabilities and real-world applications, enabling benchmarks to transcend specific capability boundaries and instead integrate multiple competencies, thereby linking benchmark scores more directly to enhanced user experience.

Meeseeks currently contains a relatively limited dataset, necessitating multiple evaluation runs to obtain stable performance scores. However, this represents a continuously evolving benchmark, and the present work primarily focuses on the novel perspective of multi-turn feedback evaluation. The work has been open-sourced, and we are committed to ongoing updates, with plans to substantially expand the Meeseeks dataset in the near future.

\bibliography{custom}

\begin{thebibliography}{57}
\providecommand{\natexlab}[1]{#1}

\bibitem[{{anthropic}(2025)}]{LLM_claude_sonnet4}
{anthropic}. 2025.
\newblock \href {https://www.anthropic.com/claude/sonnet} {Claude sonnet 4}.
\newblock Visited date: 2025-09-15.

\bibitem[{Arun et~al.(2023)Arun, Dhiman, Soni, and Hu}]{arun2023numericalreasoningfinancialreports}
Abhinav Arun, Ashish Dhiman, Mehul Soni, and Yibei Hu. 2023.
\newblock \href {https://arxiv.org/abs/2312.14870} {Numerical reasoning for financial reports}.
\newblock \emph{Preprint}, arXiv:2312.14870.

\bibitem[{Bang et~al.(2025)Bang, Ji, Schelten, Hartshorn, Fowler, Zhang, Cancedda, and Fung}]{bang2025hallulens}
Yejin Bang, Ziwei Ji, Alan Schelten, Anthony Hartshorn, Tara Fowler, Cheng Zhang, Nicola Cancedda, and Pascale Fung. 2025.
\newblock \href {https://arxiv.org/abs/2504.17550} {Hallulens: Llm hallucination benchmark}.
\newblock \emph{Preprint}, arXiv:2504.17550.

\bibitem[{{Bytedance}(2025)}]{LLM_doubao2024}
{Bytedance}. 2025.
\newblock \href {https://www.doubao.com/chat/} {Doubao-ai}.
\newblock Visited date: 2025-09-15.

\bibitem[{Chen et~al.(2024)Chen, Chen, Zhou, He, and He}]{chen2024diahalu}
Kedi Chen, Qin Chen, Jie Zhou, Yishen He, and Liang He. 2024.
\newblock \href {https://arxiv.org/abs/2403.00896} {Diahalu: A dialogue-level hallucination evaluation benchmark for large language models}.
\newblock \emph{Preprint}, arXiv:2403.00896.

\bibitem[{DeepSeek-AI(2024)}]{LLM_dsv3}
DeepSeek-AI. 2024.
\newblock \href {https://arxiv.org/abs/2412.19437} {Deepseek-v3 technical report}.
\newblock \emph{Preprint}, arXiv:2412.19437.

\bibitem[{Ganguli et~al.(2023)Ganguli, Askell, Schiefer, Liao, Lukošiūtė, Chen, Goldie, Mirhoseini, Olsson, Hernandez, Drain, Li, Tran-Johnson, Perez, Kernion, Kerr, Mueller, Landau, Ndousse, Nguyen, Lovitt, Sellitto, Elhage, Mercado, DasSarma, Rausch, Lasenby, Larson, Ringer, Kundu, Kadavath, Johnston, Kravec, Showk, Lanham, Telleen-Lawton, Henighan, Hume, Bai, Hatfield-Dodds, Mann, Amodei, Joseph, McCandlish, Brown, Olah, Clark, Bowman, and Kaplan}]{schiefer2023capacity}
Deep Ganguli, Amanda Askell, Nicholas Schiefer, Thomas~I. Liao, Kamilė Lukošiūtė, Anna Chen, Anna Goldie, Azalia Mirhoseini, Catherine Olsson, Danny Hernandez, Dawn Drain, Dustin Li, Eli Tran-Johnson, Ethan Perez, Jackson Kernion, Jamie Kerr, Jared Mueller, Joshua Landau, Kamal Ndousse, and 30 others. 2023.
\newblock \href {https://arxiv.org/abs/2302.07459} {The capacity for moral self-correction in large language models}.
\newblock \emph{Preprint}, arXiv:2302.07459.

\bibitem[{{Google DeepMind}(2025)}]{LLM_gemini_flash_2024}
{Google DeepMind}. 2025.
\newblock \href {https://deepmind.google/models/gemini/flash/} {Gemini flash}.
\newblock Visited date: 2025-09-15.

\bibitem[{Gu et~al.(2024)Gu, Yin, Liu, and Zhang}]{gu2024medvh}
Zishan Gu, Changchang Yin, Fenglin Liu, and Ping Zhang. 2024.
\newblock \href {https://arxiv.org/abs/2407.02730} {Medvh: Towards systematic evaluation of hallucination for large vision language models in the medical context}.
\newblock \emph{Preprint}, arXiv:2407.02730.

\bibitem[{Hao et~al.(2025)Hao, Yu, and You}]{Hao2025}
Yijie Hao, Haofei Yu, and Jiaxuan You. 2025.
\newblock \href {https://arxiv.org/abs/2506.06539} {Beyond facts: Evaluating intent hallucination in large language models}.
\newblock \emph{Preprint}, arXiv:2506.06539.

\bibitem[{He et~al.(2024{\natexlab{a}})He, Zeng, Huang, Chen, Xiao, He, Zhou, Chen, Wang, Huang, Ye, Li, Chen, Zhang, Gu, Liang, and Xiao}]{single_benchmark_CELLO}
Qianyu He, Jie Zeng, Wenhao Huang, Lina Chen, Jin Xiao, Qianxi He, Xunzhe Zhou, Lida Chen, Xintao Wang, Yuncheng Huang, Haoning Ye, Zihan Li, Shisong Chen, Yikai Zhang, Zhouhong Gu, Jiaqing Liang, and Yanghua Xiao. 2024{\natexlab{a}}.
\newblock \href {https://arxiv.org/abs/2309.09150} {Can large language models understand real-world complex instructions?}
\newblock \emph{Preprint}, arXiv:2309.09150.

\bibitem[{He et~al.(2024{\natexlab{b}})He, Jin, Wang, Bi, Mandyam, Zhang, Zhu, Li, Xu, Lv, Bhosale, Zhu, Sankararaman, Helenowski, Kambadur, Tayade, Ma, Fang, and Wang}]{multi_benchmark_Multi-IF}
Yun He, Di~Jin, Chaoqi Wang, Chloe Bi, Karishma Mandyam, Hejia Zhang, Chen Zhu, Ning Li, Tengyu Xu, Hongjiang Lv, Shruti Bhosale, Chenguang Zhu, Karthik~Abinav Sankararaman, Eryk Helenowski, Melanie Kambadur, Aditya Tayade, Hao Ma, Han Fang, and Sinong Wang. 2024{\natexlab{b}}.
\newblock \href {https://arxiv.org/abs/2410.15553} {Multi-if: Benchmarking llms on multi-turn and multilingual instructions following}.
\newblock \emph{Preprint}, arXiv:2410.15553.

\bibitem[{Hiraoka and Okazaki(2024)}]{word_count1}
Tatsuya Hiraoka and Naoaki Okazaki. 2024.
\newblock \href {https://arxiv.org/abs/2402.09808} {Knowledge of pretrained language models on surface information of tokens}.
\newblock \emph{Preprint}, arXiv:2402.09808.

\bibitem[{Hou et~al.(2025)Hou, Chen, Wang, Yin, Wang, Du, Pang, Chang, and Lei}]{intro2.5}
Bairu Hou, Qibin Chen, Jianyu Wang, Guoli Yin, Chong Wang, Nan Du, Ruoming Pang, Shiyu Chang, and Tao Lei. 2025.
\newblock \href {https://arxiv.org/abs/2501.02086} {Instruction-following pruning for large language models}.
\newblock \emph{Preprint}, arXiv:2501.02086.

\bibitem[{Huang et~al.(2024)Huang, Chen, Mishra, Zheng, Yu, Song, and Zhou}]{huang2023large}
Jie Huang, Xinyun Chen, Swaroop Mishra, Huaixiu~Steven Zheng, Adams~Wei Yu, Xinying Song, and Denny Zhou. 2024.
\newblock \href {https://arxiv.org/abs/2310.01798} {Large language models cannot self-correct reasoning yet}.
\newblock \emph{Preprint}, arXiv:2310.01798.

\bibitem[{Huang et~al.(2025)Huang, Yu, Ma, Zhong, Feng, Wang, Chen, Peng, Feng, Qin, and Liu}]{Huang2024}
Lei Huang, Weijiang Yu, Weitao Ma, Weihong Zhong, Zhangyin Feng, Haotian Wang, Qianglong Chen, Weihua Peng, Xiaocheng Feng, Bing Qin, and Ting Liu. 2025.
\newblock \href {https://doi.org/10.1145/3703155} {A survey on hallucination in large language models: Principles, taxonomy, challenges, and open questions}.
\newblock \emph{ACM Transactions on Information Systems}, 43(2):1–55.

\bibitem[{Jiang et~al.(2024)Jiang, Wang, Zeng, Zhong, Li, Mi, Shang, Jiang, Liu, and Wang}]{multi_benchmark:followbench}
Yuxin Jiang, Yufei Wang, Xingshan Zeng, Wanjun Zhong, Liangyou Li, Fei Mi, Lifeng Shang, Xin Jiang, Qun Liu, and Wei Wang. 2024.
\newblock \href {https://arxiv.org/abs/2310.20410} {Followbench: A multi-level fine-grained constraints following benchmark for large language models}.
\newblock \emph{Preprint}, arXiv:2310.20410.

\bibitem[{Kaplan et~al.(2025)Kaplan, Oren, Reif, and Schwartz}]{word_count3}
Guy Kaplan, Matanel Oren, Yuval Reif, and Roy Schwartz. 2025.
\newblock \href {https://arxiv.org/abs/2410.05864} {From tokens to words: On the inner lexicon of llms}.
\newblock \emph{Preprint}, arXiv:2410.05864.

\bibitem[{Kim et~al.(2025)Kim, Muhn, and Nikolaev}]{kim2025financialstatementanalysislarge}
Alex Kim, Maximilian Muhn, and Valeri Nikolaev. 2025.
\newblock \href {https://arxiv.org/abs/2407.17866} {Financial statement analysis with large language models}.
\newblock \emph{Preprint}, arXiv:2407.17866.

\bibitem[{Kumar et~al.(2024)Kumar, Zhuang, Agarwal, Su, Co-Reyes, Singh, Baumli, Iqbal, Bishop, Roelofs, Zhang, McKinney, Shrivastava, Paduraru, Tucker, Precup, Behbahani, and Faust}]{zhuang2024training}
Aviral Kumar, Vincent Zhuang, Rishabh Agarwal, Yi~Su, John~D Co-Reyes, Avi Singh, Kate Baumli, Shariq Iqbal, Colton Bishop, Rebecca Roelofs, Lei~M Zhang, Kay McKinney, Disha Shrivastava, Cosmin Paduraru, George Tucker, Doina Precup, Feryal Behbahani, and Aleksandra Faust. 2024.
\newblock \href {https://arxiv.org/abs/2409.12917} {Training language models to self-correct via reinforcement learning}.
\newblock \emph{Preprint}, arXiv:2409.12917.

\bibitem[{Lee et~al.(2023)Lee, Pujara, Sewak, White, and Jauhar}]{intro1.4}
Dong-Ho Lee, Jay Pujara, Mohit Sewak, Ryen~W. White, and Sujay~Kumar Jauhar. 2023.
\newblock \href {https://arxiv.org/abs/2310.20111} {Making large language models better data creators}.
\newblock \emph{Preprint}, arXiv:2310.20111.

\bibitem[{Li et~al.(2025{\natexlab{a}})Li, Li, Wang, Chang, and Wu}]{multi_benchmark_StructFlowBench}
Jinnan Li, Jinzhe Li, Yue Wang, Yi~Chang, and Yuan Wu. 2025{\natexlab{a}}.
\newblock \href {https://arxiv.org/abs/2502.14494} {Structflowbench: A structured flow benchmark for multi-turn instruction following}.
\newblock \emph{Preprint}, arXiv:2502.14494.

\bibitem[{Li et~al.(2023{\natexlab{a}})Li, Cheng, Zhao, Nie, and Wen}]{li2023halueval}
Junyi Li, Xiaoxue Cheng, Wayne~Xin Zhao, Jian-Yun Nie, and Ji-Rong Wen. 2023{\natexlab{a}}.
\newblock \href {https://arxiv.org/abs/2305.11747} {Halueval: A large-scale hallucination evaluation benchmark for large language models}.
\newblock \emph{Preprint}, arXiv:2305.11747.

\bibitem[{Li et~al.(2025{\natexlab{b}})Li, Yu, Zhang, Chen, Zhang, Zhuang, Sadagopan, and Beniwal}]{wang2025explicit}
Xiaomin Li, Zhou Yu, Zhiwei Zhang, Xupeng Chen, Ziji Zhang, Yingying Zhuang, Narayanan Sadagopan, and Anurag Beniwal. 2025{\natexlab{b}}.
\newblock \href {https://arxiv.org/abs/2505.11423} {When thinking fails: The pitfalls of reasoning for instruction-following in llms}.
\newblock \emph{Preprint}, arXiv:2505.11423.

\bibitem[{Li et~al.(2023{\natexlab{b}})Li, Li, Zhang, Dan, Jiang, and Zhang}]{li2023chatdoctormedicalchatmodel}
Yunxiang Li, Zihan Li, Kai Zhang, Ruilong Dan, Steve Jiang, and You Zhang. 2023{\natexlab{b}}.
\newblock \href {https://arxiv.org/abs/2303.14070} {Chatdoctor: A medical chat model fine-tuned on a large language model meta-ai (llama) using medical domain knowledge}.
\newblock \emph{Preprint}, arXiv:2303.14070.

\bibitem[{Liu et~al.(2024{\natexlab{a}})Liu, Nassereldine, Yang, Xu, Hu, Li, Kumar, Lee, Qin, Shi, and Xiong}]{liu2024large}
Dancheng Liu, Amir Nassereldine, Ziming Yang, Chenhui Xu, Yuting Hu, Jiajie Li, Utkarsh Kumar, Changjae Lee, Ruiyang Qin, Yiyu Shi, and Jinjun Xiong. 2024{\natexlab{a}}.
\newblock \href {https://arxiv.org/abs/2406.15673} {Large language models have intrinsic self-correction ability}.
\newblock \emph{Preprint}, arXiv:2406.15673.

\bibitem[{Liu et~al.(2024{\natexlab{b}})Liu, Mao, Cao, Xue, Zhang, Wang, Tang, and Johnson}]{mao2024intrinsic}
Guangliang Liu, Haitao Mao, Bochuan Cao, Zhiyu Xue, Xitong Zhang, Rongrong Wang, Jiliang Tang, and Kristen Johnson. 2024{\natexlab{b}}.
\newblock \href {https://arxiv.org/abs/2406.02378} {On the intrinsic self-correction capability of llms: Uncertainty and latent concept}.
\newblock \emph{Preprint}, arXiv:2406.02378.

\bibitem[{Minaee et~al.(2025)Minaee, Mikolov, Nikzad, Chenaghlu, Socher, Amatriain, and Gao}]{intro1.1}
Shervin Minaee, Tomas Mikolov, Narjes Nikzad, Meysam Chenaghlu, Richard Socher, Xavier Amatriain, and Jianfeng Gao. 2025.
\newblock \href {https://arxiv.org/abs/2402.06196} {Large language models: A survey}.
\newblock \emph{Preprint}, arXiv:2402.06196.

\bibitem[{{OpenAI}(2025{\natexlab{a}})}]{LLM_gpt4_1}
{OpenAI}. 2025{\natexlab{a}}.
\newblock \href {https://openai.com/index/gpt-4-1/} {Gpt-4.1}.
\newblock Visited date: 2025-09-15.

\bibitem[{{OpenAI}(2025{\natexlab{b}})}]{LLM_o3mini}
{OpenAI}. 2025{\natexlab{b}}.
\newblock \href {https://openai.com/zh-Hans-CN/index/openai-o3-mini/} {o3-mini}.
\newblock Visited date: 2025-09-15.

\bibitem[{Ouyang et~al.(2022)Ouyang, Wu, Jiang, Almeida, Wainwright, Mishkin, Zhang, Agarwal, Slama, Ray, Schulman, Hilton, Kelton, Miller, Simens, Askell, Welinder, Christiano, Leike, and Lowe}]{intro2.1}
Long Ouyang, Jeff Wu, Xu~Jiang, Diogo Almeida, Carroll~L. Wainwright, Pamela Mishkin, Chong Zhang, Sandhini Agarwal, Katarina Slama, Alex Ray, John Schulman, Jacob Hilton, Fraser Kelton, Luke Miller, Maddie Simens, Amanda Askell, Peter Welinder, Paul Christiano, Jan Leike, and Ryan Lowe. 2022.
\newblock \href {https://arxiv.org/abs/2203.02155} {Training language models to follow instructions with human feedback}.
\newblock \emph{Preprint}, arXiv:2203.02155.

\bibitem[{Pandya and Holia(2023)}]{intro1.3}
Keivalya Pandya and Mehfuza Holia. 2023.
\newblock \href {https://arxiv.org/abs/2310.05421} {Automating customer service using langchain: Building custom open-source gpt chatbot for organizations}.
\newblock \emph{Preprint}, arXiv:2310.05421.

\bibitem[{Qian et~al.(2024)Qian, Han, Fung, Qin, Liu, and Ji}]{intro1.5}
Cheng Qian, Chi Han, Yi~R. Fung, Yujia Qin, Zhiyuan Liu, and Heng Ji. 2024.
\newblock \href {https://arxiv.org/abs/2305.14318} {Creator: Tool creation for disentangling abstract and concrete reasoning of large language models}.
\newblock \emph{Preprint}, arXiv:2305.14318.

\bibitem[{Qin et~al.(2024)Qin, Song, Hu, Yao, Cho, Wang, Wu, Liu, Liu, and Yu}]{single_benchmark:InFoBench}
Yiwei Qin, Kaiqiang Song, Yebowen Hu, Wenlin Yao, Sangwoo Cho, Xiaoyang Wang, Xuansheng Wu, Fei Liu, Pengfei Liu, and Dong Yu. 2024.
\newblock \href {https://arxiv.org/abs/2401.03601} {Infobench: Evaluating instruction following ability in large language models}.
\newblock \emph{Preprint}, arXiv:2401.03601.

\bibitem[{Qiu et~al.(2024{\natexlab{a}})Qiu, Huang, Gao, Qi, Zhang, Shao, and Lu}]{qiu2024longhalqa}
Han Qiu, Jiaxing Huang, Peng Gao, Qin Qi, Xiaoqin Zhang, Ling Shao, and Shijian Lu. 2024{\natexlab{a}}.
\newblock \href {https://arxiv.org/abs/2410.09962} {Longhalqa: Long-context hallucination evaluation for multimodal large language models}.
\newblock \emph{Preprint}, arXiv:2410.09962.

\bibitem[{Qiu et~al.(2024{\natexlab{b}})Qiu, Lam, Li et~al.}]{qiu2024llmbased}
J.~Qiu, K.~Lam, G.~Li, and 1 others. 2024{\natexlab{b}}.
\newblock \href {https://doi.org/10.1038/s42256-024-00944-1} {Llm-based agentic systems in medicine and healthcare}.
\newblock \emph{Nature Machine Intelligence}, 6(12):1418--1420.

\bibitem[{Qwen et~al.(2025)Qwen, :, Yang, Yang, Zhang, Hui, Zheng, Yu, Li, Liu, Huang, Wei, Lin, Yang, Tu, Zhang, Yang, Yang, Zhou, Lin, Dang, Lu, Bao, Yang, Yu, Li, Xue, Zhang, Zhu, Men, Lin, Li, Tang, Xia, Ren, Ren, Fan, Su, Zhang, Wan, Liu, Cui, Zhang, and Qiu}]{EVAL_LLM_qwen2_5}
Qwen, :, An~Yang, Baosong Yang, Beichen Zhang, Binyuan Hui, Bo~Zheng, Bowen Yu, Chengyuan Li, Dayiheng Liu, Fei Huang, Haoran Wei, Huan Lin, Jian Yang, Jianhong Tu, Jianwei Zhang, Jianxin Yang, Jiaxi Yang, Jingren Zhou, and 25 others. 2025.
\newblock \href {https://arxiv.org/abs/2412.15115} {Qwen2.5 technical report}.
\newblock \emph{Preprint}, arXiv:2412.15115.

\bibitem[{Stolfo et~al.(2025)Stolfo, Balachandran, Yousefi, Horvitz, and Nushi}]{intro2.4}
Alessandro Stolfo, Vidhisha Balachandran, Safoora Yousefi, Eric Horvitz, and Besmira Nushi. 2025.
\newblock \href {https://arxiv.org/abs/2410.12877} {Improving instruction-following in language models through activation steering}.
\newblock \emph{Preprint}, arXiv:2410.12877.

\bibitem[{Sun et~al.(2024)Sun, Liu, Zhou, Huang, Song, Zhao, Zhang, Zhang, and Gai}]{multi_benchmark_Parrot}
Yuchong Sun, Che Liu, Kun Zhou, Jinwen Huang, Ruihua Song, Wayne~Xin Zhao, Fuzheng Zhang, Di~Zhang, and Kun Gai. 2024.
\newblock \href {https://arxiv.org/abs/2310.07301} {Parrot: Enhancing multi-turn instruction following for large language models}.
\newblock \emph{Preprint}, arXiv:2310.07301.

\bibitem[{Team et~al.(2025{\natexlab{a}})Team, Zeng, Lv, Zheng, Hou, Chen, Xie, Wang, Yin, Zeng, Zhang, Wang, Zhong, Liu, Lu, Cao, Zhang, Huang, Wei, Cheng, An, Niu, Wen, Bai, Du, Wang, Zhu, Zhang, Wen, Wu, Xu, Huang, Zhao, Cai, Yu, Li, Ge, Huang, Zhang, Xu, Zhu, Li, Yin, Lin, Yang, Jiang, Ai, Zhu, Wang, Pan, Wang, Sun, Li, Li, Hu, Zhang, Peng, Tai, Zhang, Wang, Yang, Liu, Zhao, Liu, Yan, Liu, Chen, Li, Zhao, Ren, Jiao, Zhao, Yan, Wang, Gui, Zhao, Liu, Li, Li, Lu, Wang, Yuan, Li, Du, Du, Liu, Zhi, Gao, Wang, Yang, Xu, Fan, Wu, Ding, Wang, Zhang, Li, Xu, Zhao, Zhai, Du, Dong, Lei, Tu, Yang, Lu, Li, Li, Shuang-Li, Yang, Yi, Yu, Tian, Wang, Yu, Tam, Liang, Liu, Wang, Jia, Gu, Ling, Wang, Fan, Pan, Zhang, Zhang, Fu, Zhang, Xu, Wu, Lu, Wang, Zhou, Pan, Zhang, Wang, Li, Su, Geng, Zhu, Yang, Li, Wu, Li, Liu, Wang, Li, Zhang, Liu, Yang, Zhou, Qiao, Feng, Liu, Zhang, Wang, Yao, Wang, Liu, Chai, Li, Zhao, Chen, Zhai, Xu, Huang, Wang, Li, Dong, and Tang}]{LLM_glm}
5~Team, Aohan Zeng, Xin Lv, Qinkai Zheng, Zhenyu Hou, Bin Chen, Chengxing Xie, Cunxiang Wang, Da~Yin, Hao Zeng, Jiajie Zhang, Kedong Wang, Lucen Zhong, Mingdao Liu, Rui Lu, Shulin Cao, Xiaohan Zhang, Xuancheng Huang, Yao Wei, and 152 others. 2025{\natexlab{a}}.
\newblock \href {https://arxiv.org/abs/2508.06471} {Glm-4.5: Agentic, reasoning, and coding (arc) foundation models}.
\newblock \emph{Preprint}, arXiv:2508.06471.

\bibitem[{Team et~al.(2025{\natexlab{b}})Team, Bai, Bao, Chen, Chen, Chen, Chen, Chen, Chen, Chen, Chen, Cui, Ding, Dong, Du, Du, Du, Du, Fan, Feng, Fu, Gao, Gao, Gao, Gao, Gu, Guan, Guo, Guo, Hu, Hao, He, He, He, Hong, Hu, Hu, Huang, Huang, Huang, Jiang, Jiang, Jin, Kang, Lai, Li, Li, Li, Li, Li, Li, Li, Li, Li, Lin, Lin, Lin, Liu, Liu, Liu, Liu, Liu, Liu, Liu, Liu, Liu, Liu, Liu, Liu, Liu, Liu, Liu, Lu, Lu, Ma, Ma, Ma, Mao, Mei, Men, Miao, Pan, Peng, Qin, Qu, Shang, Shi, Shi, Song, Su, Su, Sun, Sung, Tang, Tao, Teng, Wang, Wang, Wang, Wang, Wang, Wang, Wang, Wang, Wang, Wang, Wang, Wang, Wang, Wang, Wang, Wang, Wang, Wei, Wei, Wu, Wu, Wu, Xiao, Xie, Xiong, Xu, Xu, Xu, Xu, Xu, Xu, Xu, Xu, Xu, Xu, Yan, Yan, Yang, Yang, Yang, Yang, Yang, Yao, Yao, Ye, Ye, Yin, Yu, Yuan, Yuan, Yuan, Zhan, Zhang, Zhang, Zhang, Zhang, Zhang, Zhang, Zhang, Zhang, Zhang, Zhang, Zhang, Zhao, Zhao, Zheng, Zheng, Zhou, Zhou, Zhou, Zhu, Zhuang, and Zu}]{LLM_kimik2}
Kimi Team, Yifan Bai, Yiping Bao, Guanduo Chen, Jiahao Chen, Ningxin Chen, Ruijue Chen, Yanru Chen, Yuankun Chen, Yutian Chen, Zhuofu Chen, Jialei Cui, Hao Ding, Mengnan Dong, Angang Du, Chenzhuang Du, Dikang Du, Yulun Du, Yu~Fan, and 150 others. 2025{\natexlab{b}}.
\newblock \href {https://arxiv.org/abs/2507.20534} {Kimi k2: Open agentic intelligence}.
\newblock \emph{Preprint}, arXiv:2507.20534.

\bibitem[{Vieira et~al.(2025)Vieira, LeBrun, Giulianelli, Gastaldi, DuSell, Terilla, O'Donnell, and Cotterell}]{word_count4}
Tim Vieira, Ben LeBrun, Mario Giulianelli, Juan~Luis Gastaldi, Brian DuSell, John Terilla, Timothy~J. O'Donnell, and Ryan Cotterell. 2025.
\newblock \href {https://arxiv.org/abs/2412.03719} {From language models over tokens to language models over characters}.
\newblock \emph{Preprint}, arXiv:2412.03719.

\bibitem[{Wang et~al.(2023)Wang, Min, Deng, Shen, Wu, Zettlemoyer, and Sun}]{wang2022towards}
Boshi Wang, Sewon Min, Xiang Deng, Jiaming Shen, You Wu, Luke Zettlemoyer, and Huan Sun. 2023.
\newblock \href {https://arxiv.org/abs/2212.10001} {Towards understanding chain-of-thought prompting: An empirical study of what matters}.
\newblock \emph{Preprint}, arXiv:2212.10001.

\bibitem[{Wei et~al.(2022)Wei, Wang, Schuurmans, Bosma, Ichter, Xia, Chi, Le, and Zhou}]{wei2022chain}
Jason Wei, Xuezhi Wang, Dale Schuurmans, Maarten Bosma, Brian Ichter, Fei Xia, Ed~Chi, Quoc Le, and Denny Zhou. 2022.
\newblock \href {https://arxiv.org/abs/2201.11903} {Chain-of-thought prompting elicits reasoning in large language models}.
\newblock \emph{Preprint}, arXiv:2201.11903.

\bibitem[{Wen et~al.(2024)Wen, Ke, Gu, Wu, Huang, Zhou, Li, Hu, Gao, Xu, Liu, Tang, Wang, and Huang}]{single_benchmark_Complexbench}
Bosi Wen, Pei Ke, Xiaotao Gu, Lindong Wu, Hao Huang, Jinfeng Zhou, Wenchuang Li, Binxin Hu, Wendy Gao, Jiaxin Xu, Yiming Liu, Jie Tang, Hongning Wang, and Minlie Huang. 2024.
\newblock \href {https://arxiv.org/abs/2407.03978} {Benchmarking complex instruction-following with multiple constraints composition}.
\newblock \emph{Preprint}, arXiv:2407.03978.

\bibitem[{Wulf and Meierhofer(2024)}]{intro1.2}
Jochen Wulf and Juerg Meierhofer. 2024.
\newblock \href {https://arxiv.org/abs/2405.09161} {Exploring the potential of large language models for automation in technical customer service}.
\newblock \emph{Preprint}, arXiv:2405.09161.

\bibitem[{Yan et~al.(2024)Yan, Zhang, Yuan, Shan, and Chen}]{yan2024evaluating}
Bei Yan, Jie Zhang, Zheng Yuan, Shiguang Shan, and Xilin Chen. 2024.
\newblock \href {https://arxiv.org/abs/2406.17115} {Evaluating the quality of hallucination benchmarks for large vision-language models}.
\newblock \emph{Preprint}, arXiv:2406.17115.

\bibitem[{Yang et~al.(2025)Yang, Li, Yang, Zhang, Hui, Zheng, Yu, Gao, Huang, Lv, Zheng, Liu, Zhou, Huang, Hu, Ge, Wei, Lin, Tang, Yang, Tu, Zhang, Yang, Yang, Zhou, Zhou, Lin, Dang, Bao, Yang, Yu, Deng, Li, Xue, Li, Zhang, Wang, Zhu, Men, Gao, Liu, Luo, Li, Tang, Yin, Ren, Wang, Zhang, Ren, Fan, Su, Zhang, Zhang, Wan, Liu, Wang, Cui, Zhang, Zhou, and Qiu}]{LLM_qwen3}
An~Yang, Anfeng Li, Baosong Yang, Beichen Zhang, Binyuan Hui, Bo~Zheng, Bowen Yu, Chang Gao, Chengen Huang, Chenxu Lv, Chujie Zheng, Dayiheng Liu, Fan Zhou, Fei Huang, Feng Hu, Hao Ge, Haoran Wei, Huan Lin, Jialong Tang, and 41 others. 2025.
\newblock \href {https://arxiv.org/abs/2505.09388} {Qwen3 technical report}.
\newblock \emph{Preprint}, arXiv:2505.09388.

\bibitem[{Yao et~al.(2023)Yao, Chen, Hanjie, Yang, and Narasimhan}]{single_benchmark_COLLIE}
Shunyu Yao, Howard Chen, Austin~W. Hanjie, Runzhe Yang, and Karthik Narasimhan. 2023.
\newblock \href {https://arxiv.org/abs/2307.08689} {Collie: Systematic construction of constrained text generation tasks}.
\newblock \emph{Preprint}, arXiv:2307.08689.

\bibitem[{Yu et~al.(2023)Yu, He, Wu, Dai, and Chen}]{yu2023towards}
Zihan Yu, Liang He, Zhen Wu, Xinyu Dai, and Jiajun Chen. 2023.
\newblock \href {https://arxiv.org/abs/2310.04959} {Towards better chain-of-thought prompting strategies: A survey}.
\newblock \emph{Preprint}, arXiv:2310.04959.

\bibitem[{Zeng et~al.(2024)Zeng, Yu, Gao, Meng, Goyal, and Chen}]{intro2.3}
Zhiyuan Zeng, Jiatong Yu, Tianyu Gao, Yu~Meng, Tanya Goyal, and Danqi Chen. 2024.
\newblock \href {https://arxiv.org/abs/2310.07641} {Evaluating large language models at evaluating instruction following}.
\newblock \emph{Preprint}, arXiv:2310.07641.

\bibitem[{Zhang et~al.(2024)Zhang, Dong, Li, Zhang, Sun, Wang, Li, Hu, Zhang, Wu, and Wang}]{intro2.8}
Shengyu Zhang, Linfeng Dong, Xiaoya Li, Sen Zhang, Xiaofei Sun, Shuhe Wang, Jiwei Li, Runyi Hu, Tianwei Zhang, Fei Wu, and Guoyin Wang. 2024.
\newblock \href {https://arxiv.org/abs/2308.10792} {Instruction tuning for large language models: A survey}.
\newblock \emph{Preprint}, arXiv:2308.10792.

\bibitem[{Zhang et~al.(2022)Zhang, Zhang, Li, and Smola}]{zhang2022automatic}
Zhuosheng Zhang, Aston Zhang, Mu~Li, and Alex Smola. 2022.
\newblock \href {https://arxiv.org/abs/2210.03493} {Automatic chain of thought prompting in large language models}.
\newblock \emph{Preprint}, arXiv:2210.03493.

\bibitem[{Zhao et~al.(2025{\natexlab{a}})Zhao, Wang, Huang, Lu, and Liu}]{word_count2}
Chenzhuo Zhao, Xinda Wang, Yue Huang, Junting Lu, and Ziqian Liu. 2025{\natexlab{a}}.
\newblock \href {https://arxiv.org/abs/2508.05468} {Tase: Token awareness and structured evaluation for multilingual language models}.
\newblock \emph{Preprint}, arXiv:2508.05468.

\bibitem[{Zhao et~al.(2025{\natexlab{b}})Zhao, Zhou, Li, Tang, Wang, Hou, Min, Zhang, Zhang, Dong, Du, Yang, Chen, Chen, Jiang, Ren, Li, Tang, Liu, Liu, Nie, and Wen}]{intro2.7}
Wayne~Xin Zhao, Kun Zhou, Junyi Li, Tianyi Tang, Xiaolei Wang, Yupeng Hou, Yingqian Min, Beichen Zhang, Junjie Zhang, Zican Dong, Yifan Du, Chen Yang, Yushuo Chen, Zhipeng Chen, Jinhao Jiang, Ruiyang Ren, Yifan Li, Xinyu Tang, Zikang Liu, and 3 others. 2025{\natexlab{b}}.
\newblock \href {https://arxiv.org/abs/2303.18223} {A survey of large language models}.
\newblock \emph{Preprint}, arXiv:2303.18223.

\bibitem[{Zhou et~al.(2023)Zhou, Lu, Mishra, Brahma, Basu, Luan, Zhou, and Hou}]{single_benchmark_IFEVAL}
Jeffrey Zhou, Tianjian Lu, Swaroop Mishra, Siddhartha Brahma, Sujoy Basu, Yi~Luan, Denny Zhou, and Le~Hou. 2023.
\newblock \href {https://arxiv.org/abs/2311.07911} {Instruction-following evaluation for large language models}.
\newblock \emph{Preprint}, arXiv:2311.07911.

\bibitem[{Zhu et~al.(2024)Zhu, Yang, and Sun}]{zhu2024haluevalwild}
Zhiying Zhu, Yiming Yang, and Zhiqing Sun. 2024.
\newblock \href {https://arxiv.org/abs/2403.04307} {Halueval-wild: Evaluating hallucinations of language models in the wild}.
\newblock \emph{Preprint}, arXiv:2403.04307.

\end{thebibliography}

\appendix

\section{Supplementary Materials}
\subsection{Ethics Statement}
Our method and algorithm do not involve any adversarial attack, and will not endanger human security. All our experiments are performed in the simulation environment, which does not involve ethical and fair issues.

\subsection{The Use of LLM}
We used a large language model as a general-purpose assistant for text editing, including grammar correction, wording and tone adjustments, punctuation, and stylistic consistency. We used llm auto-fill function to accelerate our coding progress when developing the evaluation pipeline.

\subsection{Reproducibility Statement}
The source code of this paper is available at \url{https://anonymous.4open.science/r/Meeseeks-156A}. We specify all the implementation details of our methods in Section~\ref{sec:expers}.

\subsection{Artifacts use}
The paper follows all corresponding open-sourced/commercial license LLM used in this paper. All existing artifacts in the paper was consistent with their intended use.
\subsection{Data source}
Our dataset do not contain and personal identifying info or offensive content. We hire professional annotators check out the data completely and find no such info. All data were collected and revised by our internal business department, who have confirmed their willingness to provide the data and granted permission for open-source publication and academic submission, and the data collection protocol follows ACL Guidelines for Ethics Reviewing.
\subsection{Running Budget}
The evaluation framework can be executed without requiring local GPU resources by utilizing commercial APIs as an alternative. For researchers with access to sufficient computational resources, we recommend deploying Qwen2.5-32B-Instruct and Qwen2.5-32B-Coder models, which necessitate approximately 80 GB of GPU memory for optimal performance. The computational time required for a complete execution is approximately 45 minutes.

\subsection{Recruitment And Payment} 
Our annotators are employed under contractual agreements, with compensation details maintained under strict confidentiality. We ensure that all compensation has been fully disbursed to all participating annotators. The annotators are all professional annotators from China and all hold bachelor degree.

\section{Complete Evaluation Result}
\label{sec:complete_experiments_results}
\begin{figure*}[!htb]
    \centering
    \includegraphics[width=1\textwidth]{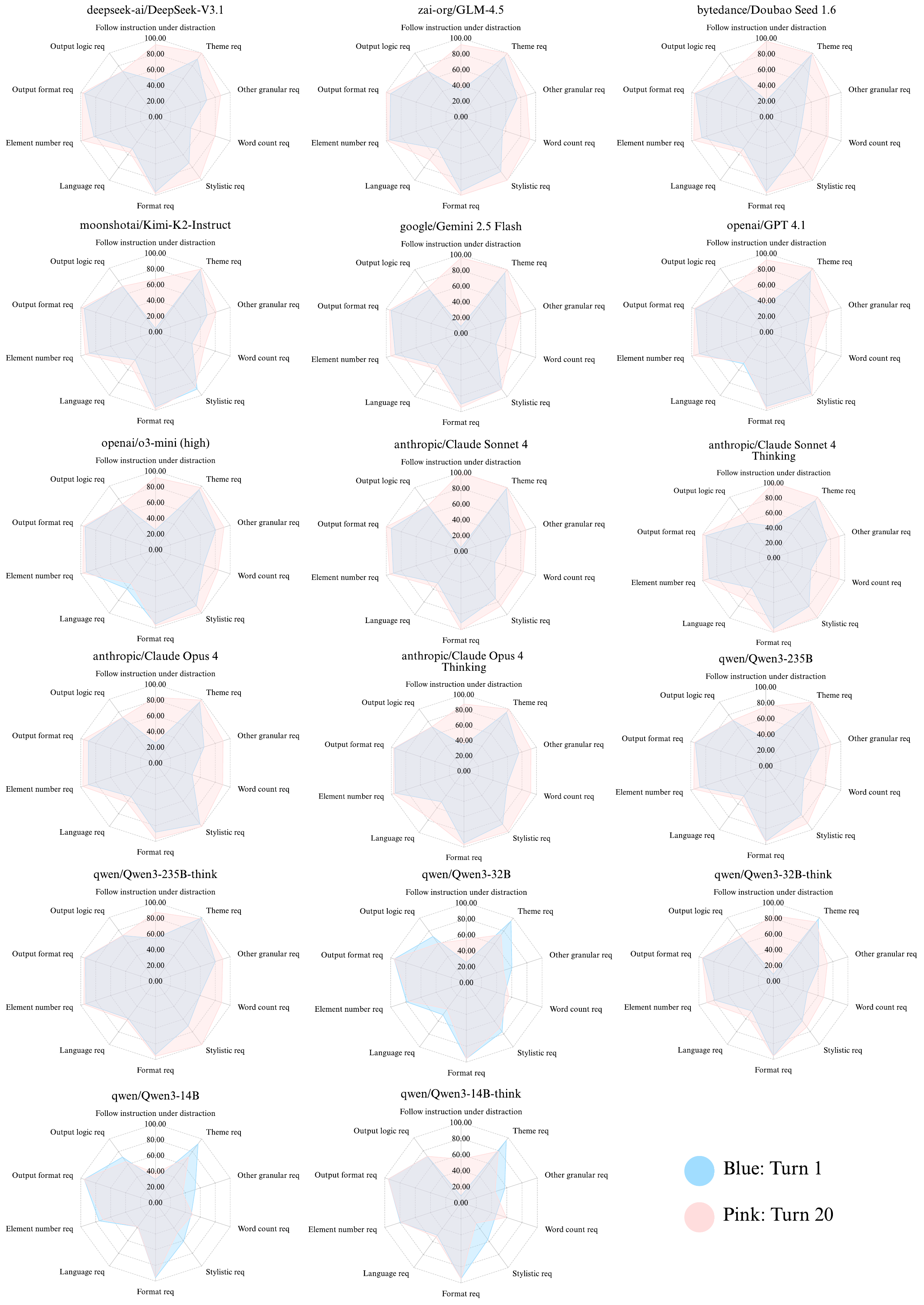}
    \caption{Models exhibit varying levels of self-correction capabilities across different dimensions, consequently demonstrating differential degrees of improvement in distinct areas.}
    \label{fig:radars}
\end{figure*}

Due to space constraints, the experimental results presented in the main text are abbreviated. Comprehensive results are provided in Table~\hyperref[tab:utility_1_10]{4} and Table~\hyperref[tab:utility_11_20]{5}. Radar graphic display the model performance on each capability tag is shown in Figure~\ref{fig:radars}. 

\begin{table*}[!htbp]
\centering
\begin{threeparttable}
\small
\begin{tabular}{>{\columncolor{white}}l r r r r r r r r r r}
\toprule
\textbf{Model ↓ \quad \quad Turn index →} & \textbf{1} & \textbf{2} & \textbf{3} & \textbf{4} & \textbf{5} & \textbf{6} & \textbf{7} & \textbf{8} & \textbf{9} & \textbf{10} \\
\midrule
\midrule
\multicolumn{1}{>{\columncolor{deepseek}}l}{deepseek-ai/DeepSeek-V3.1*} & .371 & .52 & .602 & .654 & .687 & .721 & .729 & .746 & .756 & .771 \\
\multicolumn{1}{>{\columncolor{zai}}l}{zai-org/GLM-4.5*} & .455 & .602 & .662 & .724 & .741 & .784 & .796 & .806 & .816 & .823 \\
\multicolumn{1}{>{\columncolor{bytedance}}l}{bytedance/Doubao Seed 1.6} & .338 & .475 & .572 & .627 & .644 & .664 & .692 & .701 & .706 & .714 \\
\multicolumn{1}{>{\columncolor{moonshotai}}l}{moonshotai/Kimi-K2-Instruct*} & .425 & .483 & .532 & .55 & .562 & .572 & .585 & .600 & .609 & .609 \\
\multicolumn{1}{>{\columncolor{google}}l}{google/Gemini 2.5 Flash} & .358 & .507 & .557 & .587 & .6 & .612 & .629 & .637 & .644 & .659 \\
\multicolumn{1}{>{\columncolor{openai}}l}{openai/GPT 4.1} & .425 & .55 & .572 & .592 & .609 & .617 & .624 & .629 & .644 & .657 \\
\multicolumn{1}{>{\columncolor{openai}}l}{openai/o3-mini (high)} & .567$^\dagger$ & .694$^\dagger$ & .744$^\dagger$ & .776$^\dagger$ & .794$^\dagger$ & .803$^\dagger$ & .811 & .816 & .816 & .818 \\
\multicolumn{1}{>{\columncolor{anthropic}}l}{anthropic/Claude Sonnet 4} & .351 & .555 & .637 & .674 & .704 & .719 & .731 & .749 & .764 & .774 \\
\multicolumn{1}{>{\columncolor{anthropic}}l}{anthropic/Claude Sonnet 4 Thinking} & .438 & .617 & .701 & .744 & .779 & .801 & .828$^\dagger$ & .833$^\dagger$ & .843$^\dagger$ & .856$^\dagger$ \\
\multicolumn{1}{>{\columncolor{anthropic}}l}{anthropic/Claude Opus 4} & .408 & .58 & .662 & .719 & .741 & .769 & .781 & .789 & .794 & .803 \\
\multicolumn{1}{>{\columncolor{anthropic}}l}{anthropic/Claude Opus 4 Thinking} & .49 & .637 & .697 & .754 & .771 & .789 & .801 & .821 & .828 & .833 \\
\multicolumn{1}{>{\columncolor{qwen}}l}{qwen/Qwen3-235B-A22B*} & .386 & .515 & .614 & .652 & .669 & .674 & .692 & .694 & .709 & .716 \\
\multicolumn{1}{>{\columncolor{qwen}}l}{qwen/Qwen3-235B-A22B-think*} & .520 & .619 & .682 & .734 & .761 & .803 & .816 & .833 & .836 & .843 \\
\multicolumn{1}{>{\columncolor{qwen}}l}{qwen/Qwen3-32B*} & .398 & .468 & .495 & .51 & .525 & .607 & .607 & .619 & .627 & .627 \\
\multicolumn{1}{>{\columncolor{qwen}}l}{qwen/Qwen3-32B-think*} & .331 & .473 & .527 & .57 & .6 & .530 & .545 & .552 & .557 & .565 \\
\multicolumn{1}{>{\columncolor{qwen}}l}{qwen/Qwen3-14B*} & .348 & .413 & .433 & .445 & .453 & .455 & .463 & .463 & .465 & .465 \\
\multicolumn{1}{>{\columncolor{qwen}}l}{qwen/Qwen3-14B-think*} & .371 & .44 & .483 & .507 & .53 & .540 & .545 & .552 & .552 & .557 \\
\bottomrule
\end{tabular}
\label{tab:utility_1_10}
\caption{Utility rate from turn 1 - 10}
\begin{tablenotes}
\footnotesize
\item[*] All values omit leading zero. Models marked with * are open-source.
\item[$\dagger$] Indicates the highest score in each turn.
\end{tablenotes}

\end{threeparttable}
\end{table*}

\begin{table*}[!htbp]
\centering
\begin{threeparttable}
\small
\begin{tabular}{>{\columncolor{white}}l r r r r r r r r r r}
\toprule
\textbf{Model ↓ \quad \quad Turn index →} & \textbf{11} & \textbf{12} & \textbf{13} & \textbf{14} & \textbf{15} & \textbf{16} & \textbf{17} & \textbf{18} & \textbf{19} & \textbf{20} \\
\midrule
\midrule
\multicolumn{1}{>{\columncolor{deepseek}}l}{deepseek-ai/DeepSeek-V3.1*} & .776 & .784 & .784 & .784 & .786 & .789 & .789 & .789 & .789 & .791 \\
\multicolumn{1}{>{\columncolor{zai}}l}{zai-org/GLM-4.5*} & .836 & .838 & .841 & .843 & .843 & .851 & .851 & .853 & .856 & .861 \\
\multicolumn{1}{>{\columncolor{bytedance}}l}{bytedance/Doubao Seed 1.6} & .721 & .736 & .746 & .754 & .764 & .764 & .764 & .766 & .766 & .769 \\
\multicolumn{1}{>{\columncolor{moonshotai}}l}{moonshotai/Kimi-K2-Instruct*} & .617 & .619 & .619 & .619 & .619 & .619 & .624 & .624 & .624 & .624 \\
\multicolumn{1}{>{\columncolor{google}}l}{google/Gemini 2.5 Flash} & .664 & .667 & .669 & .669 & .669 & .669 & .672 & .672 & .672 & .672 \\
\multicolumn{1}{>{\columncolor{openai}}l}{openai/GPT 4.1} & .662 & .667 & .674 & .679 & .682 & .684 & .687 & .689 & .689 & .694 \\
\multicolumn{1}{>{\columncolor{openai}}l}{openai/o3-mini (high)} & .823 & .828 & .831 & .836 & .836 & .838 & .841 & .843 & .843 & .846 \\
\multicolumn{1}{>{\columncolor{anthropic}}l}{anthropic/Claude Sonnet 4} & .779 & .784 & .784 & .784 & .784 & .789 & .789 & .791 & .794 & .794 \\
\multicolumn{1}{>{\columncolor{anthropic}}l}{anthropic/Claude Sonnet 4 Thinking} & .866$^\dagger$ & .871$^\dagger$ & .881$^\dagger$ & .891$^\dagger$ & .893$^\dagger$ & .896$^\dagger$ & .898$^\dagger$ & .898$^\dagger$ & .898$^\dagger$ & .905$^\dagger$ \\
\multicolumn{1}{>{\columncolor{anthropic}}l}{anthropic/Claude Opus 4} & .813 & .816 & .821 & .828 & .833 & .833 & .833 & .838 & .873 & .876 \\
\multicolumn{1}{>{\columncolor{anthropic}}l}{anthropic/Claude Opus 4 Thinking} & .841 & .858 & .861 & .863 & .868 & .868 & .868 & .868 & .873 & .876 \\
\multicolumn{1}{>{\columncolor{qwen}}l}{qwen/Qwen3-235B-A22B*} & .721 & .726 & .729 & .729 & .729 & .731 & .734 & .734 & .734 & .739 \\
\multicolumn{1}{>{\columncolor{qwen}}l}{qwen/Qwen3-235B-A22B-think*} & .843 & .848 & .853 & .858 & .858 & .861 & .866 & .866 & .868 & .871 \\
\multicolumn{1}{>{\columncolor{qwen}}l}{qwen/Qwen3-32B*} & .629 & .637 & .647 & .649 & .649 & .649 & .652 & .657 & .662 & .664 \\
\multicolumn{1}{>{\columncolor{qwen}}l}{qwen/Qwen3-32B-think*} & .565 & .565 & .565 & .565 & .565 & .565 & .570 & .570 & .572 & .572 \\
\multicolumn{1}{>{\columncolor{qwen}}l}{qwen/Qwen3-14B*} & .468 & .470 & .470 & .470 & .470 & .470 & .470 & .470 & .470 & .470 \\
\multicolumn{1}{>{\columncolor{qwen}}l}{qwen/Qwen3-14B-think*} & .565 & .567 & .567 & .567 & .567 & .570 & .570 & .572 & .575 & .580 \\
\bottomrule
\end{tabular}
\caption{Utility rate from turn 11 - 20}
\label{tab:utility_11_20}

\begin{tablenotes}
\footnotesize
\item[*] All values omit leading zero. Models marked with * are open-source.
\item[$\dagger$] Indicates the highest score in each turn.
\end{tablenotes}

\end{threeparttable}
\end{table*}

\section{Metric}
\label{sec:metrics_formula}
\subsection{Utility Rate}
To evaluate LLMs as trustworthy agents, we adopt and generalize IF-Eval \cite{single_benchmark_IFEVAL} 's strict "is followed" metric, introducing it as Utility Rate - the proportion of responses that fully satisfy all prompt requirements. This intuitive metric measures the ratio of truly usable responses, serving as a crucial indicator of an LLM's capability to function as a reliable professional agent. 
{\small
\begin{equation*}
\text{Utility Rate} = \frac{\sum_{i=1}^{n} U_i}{n}
\end{equation*}
\begin{equation*}
U_i = 
\begin{cases}
1, & \text{if response is usable} \\
0, & \text{if response is not usable}
\end{cases}
\end{equation*}
}
\subsection{Meeseeks Score} 
To help researchers identify subtle differences in instruction-following abilities between different LLMs, we propose Meeseeks Score, which indicates the overall capability tag accuracy of the LLM.
Meeseeks Score averages the scores of all the level 1 capability tags associated with the user prompt. 
{\small
\begin{equation*}
\text{Meeseeks Score} = \frac{\sum_{j=1}^{m} \text{Score}_{\text{tag}_j}}{m}
\end{equation*}
}
where \(m\) is the total number of level 1 capability tags associated with the current issue.

The score for each individual level 1 capability tag is the average of the scores of its respective requirements.
{\small
\begin{equation*}
\text{Score}_{\text{tag}} = \frac{\sum_{i=1}^{n} \text{Score}_{\text{requirement}_i}}{n}
\end{equation*}
}
where \(n\) is the total number of requirements for the capability tag.

\section{Data Structure}
\label{sec:data_structure}
\begin{figure*}
\centering
\includegraphics[height=8cm]{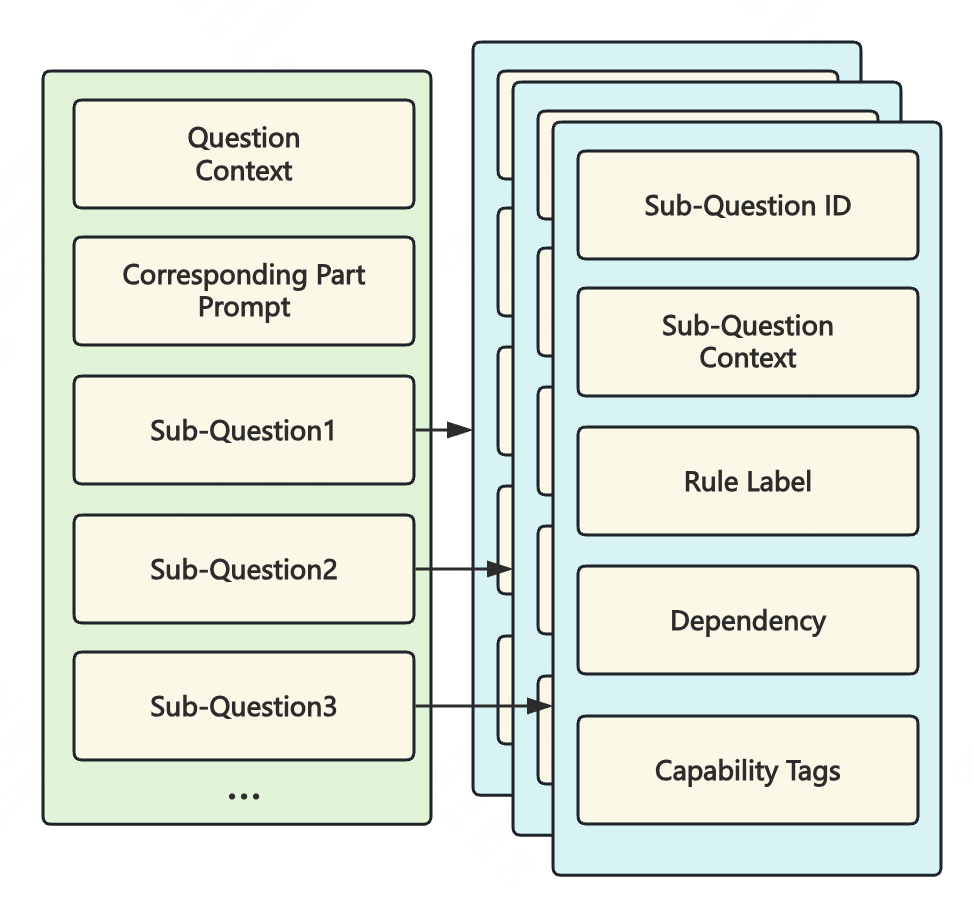}
\caption{data-structure}
\label{fig:fig4}
\end{figure*}

In this section, we delineate the composition of a singular data, providing a comprehensive overview of its constituent elements. The general arrangement and interrelations of all components within a data are visually represented in Figure \ref{fig:fig4} and a sample data is provided in Table \ref{tab:comments_generation}. 

\section{Capability Tags}
\label{sec:appendixB}
Meeseeks is equipped with a systematic evaluation system. Each sub-question is mounted with one or more capability tags, which represent the most granular level of evaluation. All capability tags are enumerated in Table \ref{tab:capability-tags-detail}, with the count of capability tags indicated in parentheses following each tag. Detailed explanations and definitions for each capability tag will be provided on GitHub, ensuring clarity and accessibility for researchers utilizing this system for evaluating model capabilities. 

Unlike existing tag-based evaluations (Table \ref{tab:system_comparison}), Meeseeks establishes a direct mapping between instruction-following abilities and benchmark performance through this systematic framework. 

Table~\ref{tab:capability-tags-detail} includes all general constraints in Chinese and English. Special language constraints will soon be released in our next project.

\begin{table*}[htbp]
\centering
\setlength{\tabcolsep}{4pt} 
\renewcommand{\arraystretch}{1.3} 
\footnotesize
\begin{tabular}{
>{\centering\arraybackslash\bfseries}m{2.8cm}
@{\hspace{0.2cm}}
>{\raggedright\arraybackslash}m{3.2cm}
@{\hspace{0.2cm}}
>{\raggedright\arraybackslash}m{7.5cm}
}
\toprule
\textbf{Level 3} & \textbf{Level 2} & \textbf{Level 1} \\
\midrule
{\begin{CJK*}{UTF8}{gbsn}任务意图理解: Intent Recognition (10)\end{CJK*}} & {\begin{CJK*}{UTF8}{gbsn}在干扰下完成指令: Follow instruction under distraction (10)\end{CJK*}} & \\[0.3em]
\midrule
\multirow{16}{=}{\centering {\begin{CJK*}{UTF8}{gbsn}单元细节合规: Granular Content Validation (2718)\end{CJK*}}} & {\begin{CJK*}{UTF8}{gbsn}主题约束: Theme requirement (922)\end{CJK*}} & \\[0.3em]
\cmidrule{2-3}
 & \multirow{2}{=}{{\begin{CJK*}{UTF8}{gbsn}文体约束: Stylistic requirement (29)\end{CJK*}}} & {\begin{CJK*}{UTF8}{gbsn}生成特定文案: Generate in certain style (17)\end{CJK*}} \\[0.2em]
 & & {\begin{CJK*}{UTF8}{gbsn}生成名字/标题: Generate names/titles (12)\end{CJK*}} \\[0.3em]
\cmidrule{2-3}
 & \multirow{2}{=}{{\begin{CJK*}{UTF8}{gbsn}语言约束: Language requirement (8)\end{CJK*}}} & {\begin{CJK*}{UTF8}{gbsn}中英文混杂: Generate Chinese-English-mixed article (8)\end{CJK*}} \\[1em]
\cmidrule{2-3}
 & \multirow{2}{=}{{\begin{CJK*}{UTF8}{gbsn}格式约束: Format requirement (529)\end{CJK*}}} & {\begin{CJK*}{UTF8}{gbsn}特定格式: Generate in other format (479)\end{CJK*}} \\[0.2em]
 & & {\begin{CJK*}{UTF8}{gbsn}日期格式: Generate result in date-format (50)\end{CJK*}} \\[0.3em]
\cmidrule{2-3}
 & \multirow{8}{=}{{\begin{CJK*}{UTF8}{gbsn}字数约束: Word count requirement (871)\end{CJK*}}} & {\begin{CJK*}{UTF8}{gbsn}精确: Generate at accurate word number (12)\end{CJK*}} \\[0.2em]
 & & {\begin{CJK*}{UTF8}{gbsn}范围: Generate in rough/range word number (437)\end{CJK*}} \\[0.2em]
 & & {\begin{CJK*}{UTF8}{gbsn}倍数: Generate in X times word number of reference text (96)\end{CJK*}} \\[0.2em]
 & & {\begin{CJK*}{UTF8}{gbsn}0～10字: Generate in 0$\sim$10 words (49)\end{CJK*}} \\[0.2em]
 & & {\begin{CJK*}{UTF8}{gbsn}10～50字: Generate in 10$\sim$50 words (145)\end{CJK*}} \\[0.2em]
 & & {\begin{CJK*}{UTF8}{gbsn}50～200字: Generate in 50$\sim$200 words (30)\end{CJK*}} \\[0.2em]
 & & {\begin{CJK*}{UTF8}{gbsn}200字以上: Generate in above 200 words (102)\end{CJK*}} \\[0.3em]
\cmidrule{2-3}
 & \multirow{9}{=}{{\begin{CJK*}{UTF8}{gbsn}其他特殊规则: Other granular requirements (359)\end{CJK*}}} & {\begin{CJK*}{UTF8}{gbsn}押韵: Generate rhyming content (16)\end{CJK*}} \\[0.2em]
 & & {\begin{CJK*}{UTF8}{gbsn}关键词: Generate with certain keywords (34)\end{CJK*}} \\[0.2em]
 & & {\begin{CJK*}{UTF8}{gbsn}重复: Generate repeat/non-repeat content (61)\end{CJK*}} \\[0.2em]
 & & {\begin{CJK*}{UTF8}{gbsn}写作手法: Generate with certain rhetoric (40)\end{CJK*}} \\[0.2em]
 & & {\begin{CJK*}{UTF8}{gbsn}词频: Generate with certain number of word X (208)\end{CJK*}} \\[0.3em]
\midrule
\multirow{3}{=}{\centering {\begin{CJK*}{UTF8}{gbsn}整体结构合规: Output Structure Validation (1346)\end{CJK*}}} & {\begin{CJK*}{UTF8}{gbsn}模版合规: Output format requirement (1255)\end{CJK*}} & {\begin{CJK*}{UTF8}{gbsn}JSON格式: Generate in JSON format (1255)\end{CJK*}} \\[0.3em]
\cmidrule{2-3}
 & {\begin{CJK*}{UTF8}{gbsn}单元数量合规: Element number requirement (71)\end{CJK*}} & \\[0.3em]
\cmidrule{2-3}
 & {\begin{CJK*}{UTF8}{gbsn}答题逻辑合规: Output logic requirement (20)\end{CJK*}} & {\begin{CJK*}{UTF8}{gbsn}答题结构合规: Generate by certain steps (20)\end{CJK*}} \\[0.3em]
\bottomrule
\end{tabular}
\caption{Capability tag system}
\label{tab:capability-tags-detail}
\end{table*}

\begin{table*}
  \centering
  \small  
  \begin{tabular}{llll}
    \toprule
    \textbf{Meeseeks} & \textbf{Complexbench} & \textbf{CELLO} & \textbf{IF-Eval} \\
    \midrule
    Intent Recognition & Format Constraint & Task-prescribed Phrases & Keywords \\
    Granular Content Validation & Lexical Constraint & Count Limit & Language \\
    Output Structure Validation & Semantic Constraint & Input-dependent Query & Length Constraints \\
    & Utility Constraint & Answer Format & Detectable Content \\
    & & & Detectable Format... \\
    \bottomrule
  \end{tabular}
  \caption{Instruction-following benchmarks evaluation system comparison}
  \label{tab:system_comparison}
\end{table*}

\newpage
\newpage

\textbf{Question Context} is the first turn prompt to the under-evaluation LLM. 

\textbf{Corresponding Part Prompt} is an prompt guiding the LLM-Extractor to extract the to-be-evaluated part from the under-evaluation LLM's response. An example is shown below.

\textbf{Sub-Question: } The \textbf{Question Context} generally contains more than one requirements. Each requirement is corresponding to one \textbf{Sub-Question}. An example is shown below.

\begin{table*}[htbp]
    \centering
    \scriptsize
    \setlength{\tabcolsep}{3pt}
    \begin{tabular}{
        @{}
        >{\raggedright\arraybackslash}p{2.5cm}|
        >{\raggedright\arraybackslash}p{1.8cm}|
        >{\raggedright\arraybackslash}p{2.1cm}|
        >{\raggedright\arraybackslash}p{2.1cm}|
        >{\raggedright\arraybackslash}p{2.1cm}|
        >{\raggedright\arraybackslash}p{2.1cm}@{}}
        \toprule
        \textbf{Question Context} & \textbf{Sub-Question Attributions} & \textbf{Sub-Question 0} & \textbf{Sub-Question 1} & \textbf{Sub-Question 2} & \textbf{Sub-Question 3} \\
        \midrule
        \multirow{5}{*}{\parbox{2.2cm}{
            \textbf{Chinese}\\
            {\begin{CJK*}{UTF8}{gbsn}在美团外卖神券短视频的评论区，以消费者角度生成80条用户口语化评论. 每条评论为10个字，不允许重复. \\[0.5ex]
            备注:字数只计中文字符\end{CJK*}}\\[0.5ex]
            \textbf{English}\\
            Generate 80 colloquial user comments from a consumer perspective in the comments section of Meituan takeaway coupon short videos. Each comment should be 10 characters long and not repeated. Note: Only count Chinese characters
        }} 
        & \textbf{Sub-Question ID} & 0 & 1 & 2 & 3 \\
        \cmidrule{2-6}
        & \textbf{Sub-Question Context} 
        & \parbox{1.5cm}{
            \textbf{Chinese}\\
            {\begin{CJK*}{UTF8}{gbsn}生成的内容是否为美团外卖神券的评论？\end{CJK*}}\\[0.5ex]
            \textbf{English}\\
            Are the generated contents comments about Meituan takeaway coupons?
        }
        & \parbox{1.5cm}{
            \textbf{Chinese}\\
            {\begin{CJK*}{UTF8}{gbsn}是否生成了80条用户口语化评论\end{CJK*}}\\[0.5ex]
            \textbf{English}\\
            Are there 80 colloquial user comments generated?
        }
        & \parbox{1.5cm}{
            \textbf{Chinese}\\
            {\begin{CJK*}{UTF8}{gbsn}所有评论是否为10个字\end{CJK*}}\\[0.5ex]
            \textbf{English}\\
            Are all comments 10 characters long?
        }
        & \parbox{1.5cm}{
            \textbf{Chinese}\\
            {\begin{CJK*}{UTF8}{gbsn}评论是否不重复\end{CJK*}}\\[0.5ex]
            \textbf{English}\\
            Are the comments non-repetitive?
        } \\
        \cmidrule{2-6}
        & \textbf{Rule Label} & Null & \parbox{1.5cm}{item\_count: [80, 80]} & \parbox{1.5cm}{each\_length: [10,10]} & non\_repeat \\
        \cmidrule{2-6}
        & \textbf{Dependency} & [] & [0] & [0] & [0] \\
        \cmidrule{2-6}
        & \textbf{Capability Tag} 
        & \parbox{1.5cm}{{\begin{CJK*}{UTF8}{gbsn}主题约束\end{CJK*}}\\Theme requirement}
        & \parbox{1.5cm}{{\begin{CJK*}{UTF8}{gbsn}单元数量合规\end{CJK*}}\\Element number requirement}
        & \parbox{1.5cm}{{\begin{CJK*}{UTF8}{gbsn}0～10字\end{CJK*}}\\Generate in 0\textasciitilde10 words\\{\begin{CJK*}{UTF8}{gbsn}精确\end{CJK*}}\\Generate at accurate word number}
        & \parbox{1.5cm}{{\begin{CJK*}{UTF8}{gbsn}重复\end{CJK*}}\\Generate repeat/non-repeat content} \\
        \cmidrule{2-6}
        & \textbf{Corresponding Part} & Null & comments & comments & comments \\
        \bottomrule
    \end{tabular}
    \caption{Breakdown of sub-questions}
    \label{tab:comments_generation}
\end{table*}

\textbf{Sub-Question Context} is the prompt input to the LLM Evaluator to judge if the under-evaluation LLM's response meets the corresponding requirement. 

\textbf{Rule Label} is a special label projecting to different rule-based evaluation function inside the Meeseeks framework. For example, \texttt{"item\_count:[80,80]"} checks whether the number of the elements(comments in the given example) is exact 80. 

\textbf{Dependency} shows the error propagation between some of the sub-questions. For example, sub-question 2 relies on sub-question 0. If the under-evaluation LLM fails on sub-question 0, it automatically fails on the sub-question 2.
 
\textbf{Capability Tag} reflects certain capability of under-evaluation LLM. It is attached with the sub-question. This will be further explained in Appendix \ref{sec:appendixB}.

\newpage
\section{Sample Data Parameterized Template}
\label{sec:datatemplate}
Below is a template example used in our work, without understanding the Chinese embedding in the template and the complicated data architecture, researchers only need to follow the instruction to generate more data:

\begin{lstlisting}[language=Python, escapeinside={(*@}{@*)}]
"""
Data Generation Template for Text Evaluation
============================================
This script generates test data for evaluating text generation models,
specifically for scenic spot recommendation tasks.

Author: Demo Template
Version: 1.1
"""

import json
import ast
import random
import re
from typing import List, Dict, Any, Tuple

# Configuration Section
# =====================

# Word count pairs for generating different test cases
# Format: [title_word_count, content_word_count]
WORD_COUNT_PAIRS = [
    [7, 200], [10, 300], [12, 300], [13, 300], 
    [9, 280], [10, 270], [11, 200], [8, 210]
]

# Output file path - modify this to change where the generated data is saved
OUTPUT_FILE_PATH = "YOUR_OUTPUT_FILEPATH"

# Template Data Definition
# ========================

# Base template for generating test cases
# This template will be used to create multiple variations
BASE_TEMPLATE_DATA = [{
    "category": "general_single",
    "question": "Recommend the scenic spot Grand Canyon, with the title needing to be exactly ###(*@{\begin{CJK*}{UTF8}{gbsn}字数\end{CJK*}}@*)1### words, and the content around ###(*@{\begin{CJK*}{UTF8}{gbsn}字数\end{CJK*}}@*)2### words. The title should be attractive and can use trending vocabulary, hot memes, and popular expressions from social media platforms like Instagram, TikTok, Twitter, etc. It should make people have a strong desire to click. Remember, the word number only counts the English words.\nNote: Please only include English words.",
    "corresponding_parts": {
        "(*@{\begin{CJK*}{UTF8}{gbsn}标题\end{CJK*}}@*)": "(*@{\begin{CJK*}{UTF8}{gbsn}请你按照\end{CJK*}}@*)python list(*@{\begin{CJK*}{UTF8}{gbsn}的格式，抓取模型回复中，给出的title部分. 同时不要提取任何注释只提取内容（例如不要提取注释：【标题】）. 比如\end{CJK*}}@*)[\"Epic Canyon Vibes: Nature's Ultimate Flex\"]",
        "(*@{\begin{CJK*}{UTF8}{gbsn}文案\end{CJK*}}@*)": "(*@{\begin{CJK*}{UTF8}{gbsn}请你按照\end{CJK*}}@*)python list(*@{\begin{CJK*}{UTF8}{gbsn}的格式，抓取模型回复中，给出的content部分. 同时不要提取任何注释只提取内容（例如不要提取注释：【标题】）. 比如\end{CJK*}}@*)[\"The Grand Canyon isn't just a destination, it's literally the main character of America's natural wonders, and honestly, it hits different every single time you visit. This geological masterpiece in Arizona serves up the most incredible views that will have you questioning if you're still on planet Earth.\"]"
    },
    "sub_questions": [
        {
            "point_id": 0,
            "question": "(*@{\begin{CJK*}{UTF8}{gbsn}是否给出了title和content\end{CJK*}}@*)",
            "rule": None,
            "dep": [],
            "(*@{\begin{CJK*}{UTF8}{gbsn}被依赖\end{CJK*}}@*)": True
        },
        {
            "point_id": 1,
            "question": "(*@{\begin{CJK*}{UTF8}{gbsn}标题是否刚好\end{CJK*}}@*)###(*@{\begin{CJK*}{UTF8}{gbsn}字数\end{CJK*}}@*)1###(*@{\begin{CJK*}{UTF8}{gbsn}个字\end{CJK*}}@*)",
            "corresponding_part": "(*@{\begin{CJK*}{UTF8}{gbsn}标题\end{CJK*}}@*)",
            "rule": "each_length:[###(*@{\begin{CJK*}{UTF8}{gbsn}字数\end{CJK*}}@*)1###,###(*@{\begin{CJK*}{UTF8}{gbsn}字数\end{CJK*}}@*)1###]",
            "dep": [0],
            "(*@{\begin{CJK*}{UTF8}{gbsn}能力项\end{CJK*}}@*)": "###(*@{\begin{CJK*}{UTF8}{gbsn}字数\end{CJK*}}@*)1###(*@{\begin{CJK*}{UTF8}{gbsn}字、范围\end{CJK*}}@*)",
            "(*@{\begin{CJK*}{UTF8}{gbsn}被依赖\end{CJK*}}@*)": False
        },
        {
            "point_id": 2,
            "question": "(*@{\begin{CJK*}{UTF8}{gbsn}文案是否在\end{CJK*}}@*)###(*@{\begin{CJK*}{UTF8}{gbsn}字数\end{CJK*}}@*)2###(*@{\begin{CJK*}{UTF8}{gbsn}字左右\end{CJK*}}@*)",
            "corresponding_part": "(*@{\begin{CJK*}{UTF8}{gbsn}文案\end{CJK*}}@*)",
            "rule": "each_length:[###(*@{\begin{CJK*}{UTF8}{gbsn}字数\end{CJK*}}@*)2*0.9###,###(*@{\begin{CJK*}{UTF8}{gbsn}字数\end{CJK*}}@*)2*1.1###]",
            "dep": [0],
            "(*@{\begin{CJK*}{UTF8}{gbsn}能力项\end{CJK*}}@*)": "###(*@{\begin{CJK*}{UTF8}{gbsn}字数\end{CJK*}}@*)2###(*@{\begin{CJK*}{UTF8}{gbsn}字、范围\end{CJK*}}@*)",
            "(*@{\begin{CJK*}{UTF8}{gbsn}被依赖\end{CJK*}}@*)": False
        }
    ]
}]

# Utility Functions
# =================

def convert_text_to_json(text: str) -> Dict[str, Any]:
    """
    Convert string representation of dictionary back to JSON object
    
    Args:
        text (str): String representation of a dictionary
        
    Returns:
        Dict[str, Any]: Parsed dictionary object
    """
    return ast.literal_eval(text)


def calculate_mathematical_expressions(text: str) -> str:
    """
    Find and evaluate mathematical expressions in the format ###expression###
    
    Args:
        text (str): Text containing mathematical expressions
        
    Returns:
        str: Text with expressions replaced by calculated results
        
    Example:
        Input: "###10*0.9###" -> Output: "9.0"
    """
    # Regular expression to find patterns like ###A*B*C###
    pattern = r'###(.*?)###'
    
    def evaluate_single_expression(match):
        """Helper function to evaluate a single mathematical expression"""
        expression = match.group(1)
        try:
            # Calculate the expression value using eval (safe for controlled input)
            result = eval(expression)
            # Format result with at most 1 decimal place
            formatted_result = "{:.1f}".format(result)
            return formatted_result
        except Exception as e:
            # If evaluation fails, return the original expression
            print(f"Warning: Could not evaluate expression '{expression}': {e}")
            return match.group(0)
    
    # Replace all mathematical expressions with calculated results
    processed_text = re.sub(pattern, evaluate_single_expression, text)
    return processed_text


def generate_test_variations(template_item: Dict[str, Any], word_pairs: List[Tuple[int, int]]) -> List[Dict[str, Any]]:
    """
    Generate multiple test case variations from a template using different word count pairs
    
    Args:
        template_item (Dict[str, Any]): Base template item to generate variations from
        word_pairs (List[Tuple[int, int]]): List of [title_words, content_words] pairs
        
    Returns:
        List[Dict[str, Any]]: List of generated test case variations
    """
    generated_variations = []
    
    for title_word_count, content_word_count in word_pairs:
        # Convert template to string for placeholder replacement
        template_string = str(template_item)
        
        # Replace word count placeholders with actual values
        # Replace ###(*@{\begin{CJK*}{UTF8}{gbsn}字数\end{CJK*}}@*)1### with title word count
        updated_template = template_string.replace("###(*@{\begin{CJK*}{UTF8}{gbsn}字数\end{CJK*}}@*)1###", str(title_word_count))
        updated_template = updated_template.replace("(*@{\begin{CJK*}{UTF8}{gbsn}字数\end{CJK*}}@*)1", str(title_word_count))
        
        # Replace ###(*@{\begin{CJK*}{UTF8}{gbsn}字数\end{CJK*}}@*)2### with content word count  
        updated_template = updated_template.replace("###(*@{\begin{CJK*}{UTF8}{gbsn}字数\end{CJK*}}@*)2###", str(content_word_count))
        updated_template = updated_template.replace("(*@{\begin{CJK*}{UTF8}{gbsn}字数\end{CJK*}}@*)2", str(content_word_count))
        
        # Calculate any mathematical expressions in the template
        updated_template = calculate_mathematical_expressions(updated_template)
        
        # Convert back to dictionary and add to results
        variation_dict = convert_text_to_json(updated_template)
        generated_variations.append(variation_dict)

    return generated_variations


def categorize_word_count_capabilities(generated_data: List[Dict[str, Any]]) -> None:
    """
    Categorize word count capabilities into predefined ranges
    
    Args:
        generated_data (List[Dict[str, Any]]): Generated test data to categorize
        
    Modifies the input data in-place by updating capability categories.
    """
    for test_item in generated_data:
        for sub_question in test_item["sub_questions"]:
            # Check if this sub-question has capability tags
            if "(*@{\begin{CJK*}{UTF8}{gbsn}能力项\end{CJK*}}@*)" in sub_question:
                capability_text = sub_question["(*@{\begin{CJK*}{UTF8}{gbsn}能力项\end{CJK*}}@*)"]
                capability_tags = capability_text.split("(*@{\begin{CJK*}{UTF8}{gbsn}、\end{CJK*}}@*)")
                
                for tag in capability_tags:
                    # Process word count related capabilities
                    if tag.endswith("(*@{\begin{CJK*}{UTF8}{gbsn}字\end{CJK*}}@*)"):
                        try:
                            # Extract numeric value from the capability tag
                            word_count = float(tag[:-1])
                            
                            # Categorize into predefined ranges
                            if 0 <= word_count < 10:
                                replacement_category = "0~10(*@{\begin{CJK*}{UTF8}{gbsn}字\end{CJK*}}@*)"
                            elif 10 <= word_count < 50:
                                replacement_category = "10~50(*@{\begin{CJK*}{UTF8}{gbsn}字\end{CJK*}}@*)"
                            elif 50 <= word_count < 200:
                                replacement_category = "50~200(*@{\begin{CJK*}{UTF8}{gbsn}字\end{CJK*}}@*)"
                            elif word_count >= 200:
                                replacement_category = "200(*@{\begin{CJK*}{UTF8}{gbsn}字以上\end{CJK*}}@*)"
                            else:
                                continue
                            
                            # Replace the specific word count with the category
                            sub_question["(*@{\begin{CJK*}{UTF8}{gbsn}能力项\end{CJK*}}@*)"] = capability_text.replace(tag, replacement_category)
                            
                        except ValueError:
                            # Skip if word count cannot be parsed as a number
                            print(f"Warning: Could not parse word count from tag '{tag}'")
                            continue


def save_generated_data(data: List[Dict[str, Any]], output_path: str) -> None:
    """
    Save the generated test data to a JSON file
    
    Args:
        data (List[Dict[str, Any]]): Generated test data to save
        output_path (str): Path where the JSON file should be saved
    """
    try:
        with open(output_path, "w", encoding='utf-8') as output_file:
            json.dump(data, output_file, indent=4, ensure_ascii=False)
        print(f"Successfully saved {len(data)} test cases to: {output_path}")
    except Exception as e:
        print(f"Error saving data to {output_path}: {e}")


# Main Execution
# ==============

def main():
    """
    Main function to orchestrate the data generation process
    """
    print("Starting data generation process...")
    print(f"Using {len(WORD_COUNT_PAIRS)} word count pairs")
    
    # Generate test case variations from the base template
    all_generated_data = []
    
    # Process the first (and currently only) template item
    base_template = BASE_TEMPLATE_DATA[0]
    generated_variations = generate_test_variations(base_template, WORD_COUNT_PAIRS)
    all_generated_data.extend(generated_variations)
    
    # Additional templates can be processed here:
    # if len(BASE_TEMPLATE_DATA) > 1:
    #     second_template = BASE_TEMPLATE_DATA[1]
    #     more_variations = generate_test_variations(second_template, WORD_COUNT_PAIRS)
    #     all_generated_data.extend(more_variations)
    
    # Categorize the word count capabilities for better organization
    categorize_word_count_capabilities(all_generated_data)
    
    # Display generation summary
    print(f"Generated {len(all_generated_data)} test cases")
    
    # Save the generated data to file
    save_generated_data(all_generated_data, OUTPUT_FILE_PATH)
    
    print("Data generation process completed successfully!")


# Script Entry Point
# ==================

if __name__ == "__main__":
    main()
\end{lstlisting}

\section{Prompts}
To maintain conciseness, only the English prompts are presented in this paper. The Chinese version is available in the GitHub repository referenced in the abstract.
\subsection{Single Element Coding Extract Prompt}
\begin{lstlisting}[escapeinside={(*@}{@*)}]
You are a Python data processing expert. You will receive [model_response] and [target_object] as input. Please write Python code to remove non-[target_object] information from [model_response] based on the [target_object].

Step 1. Find and remove non-[target_object] information at the beginning and end
Please check if there is non-[target_object] information at the beginning or end of model_response. If it doesn't exist, skip this step; if it exists, use the re.sub method to remove unwanted non-[target_object] information from model_response to ensure the output only contains the specified content of [target_object].

Step 2. Find and remove non-[target_object] information in the middle
Please check if there is non-[target_object] information nested in the middle of model_response. If it exists, use the replace method to remove it.

Common non-[target_object] information includes: "Here is the perfect essay:", "Above is all the content, hope you are satisfied"

Step 3. Output data
Output the data from model_response in Python list format after removing non-[target_object] information.

#Output Format#:
Assuming you already have "model_response":
Your code format must strictly follow the below: ensure your function name is extract_info_list. Your return result should be a list. Please only output the function: extract_info_list(model_response)

Detailed format:
def extract_info_list(model_response):
    # 1. First check if non-[target_object] information exists at the beginning and end. If not, skip this step directly.
    # If [xxxxx] is beginning non-[target_object] info and [yyyyy] is ending non-[target_object] info, use sub to delete [xxxxx] and [yyyyy]
    # For example: cleaned_text = re.sub(r'^.*xxxxx|yyyyy.*$', '', model_response, flags=re.DOTALL)
    
    # 2. If non-[target_object] information is nested within [target_object] information, use replace method to remove non-[target_object] information
    # If non-[target_object] information is zzzzzz, remove it as follows: model_response = model_response.replace("zzzzzz", "")
    
    # In many cases, there is actually no non-[target_object] information, so Step1 and Step2 can be skipped, directly assign model_response to cleaned_text
    # For example: cleaned_text = model_response
    
    # 3. Output data after removing extra information in python list format
    return [cleaned_text]

**Please note: Only output the function: extract_info_list(model_response), only output code! Do not output any other content!**

---example1---
[model_response]
Evaluating student essay: Advantages: 1. Clear intent: The article closely follows the material theme, emphasizes the importance of cooperation and sharing, and develops arguments around this core. 2. Clear structure: The article transitions from introduction to main body to conclusion with distinct levels and clear logic. 3. Fluent language: The article uses some rhetorical techniques such as metaphor and parallelism, making it literary. 4. Wide scope: The author elaborates from three levels: national, commercial, and personal, making the content rich and multi-angled. Disadvantages: 1. Lack of innovation: Although the article has a complete structure, it lacks unique insights and deep thinking, and the content is somewhat mediocre. 2. Examples not specific enough: Although national cooperation, corporate cooperation and personal sharing are mentioned, no specific examples are given, making the argument less powerful. 3. Emotional expression slightly weak: The article is more rational analysis, but emotional rendering is insufficient, lacking expression of deep feelings about the theme. Overall score: Based on the above analysis, this essay has complete content and clear structure, but is slightly insufficient in innovation and argumentative strength. If the full score is 60 points, I would give this article about 45 points. It is a well-structured article, but needs more unique insights and specific examples to improve the score. Try writing a perfect college entrance examination essay: [Title] Radiant Light, Brilliant Because of You [Main Text] In a vast sea of flowers, each flower blooms in its unique posture, together composing the chapter of spring. As General Secretary Xi Jinping said, one person's brilliance may illuminate a corner, but only when stars surround the moon can the entire night sky be illuminated. In the long river of human history, those behaviors that try to dominate and be self-admiring are ultimately fleeting. The wheel of history rolls forward, driven by those who understand cooperation and sharing. They understand that blowing out others' lights not only fails to illuminate their own path, but makes the world darker.

[target_object]
Main text content

[extract function]
**Please note: Only output the function: extract_info_list(model_response), only output code! Do not output any other content!**
def extract_info_list(model_response):
    # Because the target object is main text content, therefore: extract content after "[Main Text]"
    cleaned_text = re.sub(r'^.*\[Main Text\]', '', model_response, flags=re.DOTALL)
    return [cleaned_text]

---example2---
[model_response]
Due to the lengthy content, the following is a simplified version of the research report outline, which you can further expand as needed: ---# Research Report: System Garbage Problems Faced by Ordinary Users and Solutions## Abstract This report investigates system garbage problems encountered by ordinary users in daily use of computers and mobile devices. It aims to understand in what scenarios users encounter system garbage problems, explore simple and effective cleaning methods, and evaluate the feasibility of developing foolproof cleaning software. ## Introduction System garbage problems refer to temporary files, cache, useless programs and file fragments generated during the use of computers and mobile devices. These garbage occupy storage space, affect device performance, and may even cause privacy leaks. Ordinary users often lack professional knowledge to effectively manage these system garbage. ## Research Methods - Online questionnaire: Distribute questionnaires to ordinary users to collect data. ---Please note, this is only a simplified version of the report outline. The actual research report needs to be written based on research data and analysis results, including detailed methodology, data analysis, user feedback and software design plans, etc., to meet the 3000-word requirement.

[target_object]
The entire research report outline.

[extract function]
**Please note: Only output the function: extract_info_list(model_response), only output code! Do not output any other content!**
def extract_info_list(model_response):
    # Because the target object is the entire research report outline, therefore: extract content after "[# Research Report:]" and before "[. ---Please note]"
    cleaned_text = re.sub(r'^.*# Research Report:|\.---Please note.*$', '', model_response, flags=re.DOTALL)
    return [cleaned_text]

---your turn---
[model_response]
{model_response}

[target_object]
{instruction}

[extract function]
**Please note: Only output the function: extract_info_list(model_response), only output code! Do not output any other content!**
\end{lstlisting}
\subsection{Multiple Element Coding Extract Prompt}
\begin{lstlisting}[escapeinside={(*@}{@*)}]
You are a Python data processing expert responsible for writing Python code to extract and format information from model_response according to different requirements.

Step 1. Remove header and footer extra remarks according to instruction
Please check if there are header or footer extra remarks in model_response. If not, skip this step. If yes, use re.sub method to remove unwanted header and footer extra remarks from model_response to ensure output only contains specified content from instruction. Use re.sub to remove header and footer extra remarks.

Step2. Remove middle extra remarks according to instruction
Please check if there are nested extra remarks in the middle of model_response. If yes, use replace method to remove them.

Common extra remarks include: [Comments below:] [Above is all content] [Note: Due to space limitations, only 40 comment examples are provided here.] [Sorry, the above content may be duplicated, please ignore duplicate content]

Step 3. Data splitting
You need to split data using re.findall according to instruction, using appropriate splitting methods (such as by index, space, punctuation, etc.) to ensure each information piece is independent. You must ensure complete extraction. For example: model_response="1. This suggestion is really good\n. 2. Strongly agree with this view.\n", should extract all information including index and content, not just index or content. Ideal result: ["1. This suggestion is really good","2. Strongly agree with this view"]. Therefore, never use code like dishes = re.findall(r'\d+\. \*\*(.*?)\*\*', model_response), but use dishes = re.findall(r'(\d+\.\s*.*?)(?=\d+\.|$)', model_response, re.DOTALL) to include content with index as output.

Output format: Assuming you already have "model_response": Your code format must strictly follow: ensure your function name is extract_info_list. Your return result should be a list. Only output function: extract_info_list(model_response), no other information.

def extract_info_list(model_response):
    # 1. First check if header and footer have extra remarks. If not, skip this step. If [xxxxx] is header extra remark and [yyyyy] is footer extra remark, use sub to delete [xxxxx] and [yyyyy]
    # Example: cleaned_text = re.sub(r'^.*xxxxx|yyyyy.*$', '', model_response, flags=re.DOTALL)
    
    # 2. If there are some middle nested extra remarks [zzzzzz] not required by instruction, use replace method to remove
    # Example: model_response = model_response.replace("zzzzzz", "")
    
    # If no extra information exists, skip Step1 and Step2, directly assign model_response to cleaned_text
    # Example: cleaned_text = model_response
    
    if cleaned_text:
        # 3. According to model response format, use regex to extract comments within matched range, capture all content including index. If original text has index, must capture corresponding index: Example: comments = re.findall(r'(\d+\.\s*.*?)(?=\d+\.|$)', cleaned_text, re.DOTALL)
        # Please fill in this part of code
        # ...
    return []

**Important: Only output function: extract_info_list(model_response), only output code! Do not output any other content!**

---example1---
#model_response#: Comments below: 1. This suggestion is absolutely fantastic 2. I totally agree with this view (I think this one is not particularly good) 3. Very insightful, worth pondering 4. This deserves a big thumbs up 5. Great suggestion, hope it gets adopted 6. I 100% agree with this view 7. You're absolutely right, I think so too 8. This idea is amazing, giving it a like 9. This view is very profound, giving it a like, above is all content, if you have any dissatisfaction, please continue our conversation.

#instruction#: Please remove non-comment information or remarks, then output comments from model_response in python list format, split output.

**Important: Only output function: extract_info_list(model_response), only output code! Do not output any other content!**

#extract function#:
def extract_info_list(model_response):
    # 1. Detected header remark [Comments below:], use "^.*below:" to remove header remark; footer extra remark [, above is all content, if you have any dissatisfaction, please continue our conversation.], use "|, above.*$" to remove footer remark. Only use few characters after header remark and few characters before footer remark to ensure regex works properly.
    cleaned_text = re.sub(r'^.*below:|, above.*$', '', model_response, flags=re.DOTALL)
    
    # 2. [(I think this one is not particularly good)] does not belong to comment content, is extra remark, use replace to remove
    cleaned_text = cleaned_text.replace("(I think this one is not particularly good)", "")
    
    if cleaned_text:
        # 3. Detected index exists in model_response, use regex to extract index+comment content
        comments = re.findall(r'(\d+\.\s*.*?)(?=\d+\.|$)', cleaned_text, re.DOTALL)
        return comments
    return []

---example2---
#model_response#: 1. The environment in the video looks very comfortable, elderly people will definitely like such a place. Above are all comments

#instruction#: Please remove non-comment information or remarks, then output comments from model_response in python list format, split output.

**Important: Only output function: extract_info_list(model_response), only output code! Do not output any other content!**

#extract function#:
def extract_info_list(model_response):
    # 1. No header remark detected; detected footer remark [Above are all comments], use "| Above.*$" to remove footer remark. Only use few characters before footer remark in sub to ensure regex works properly.
    cleaned_text = re.sub(r'| Above.*$', '', model_response, flags=re.DOTALL)
    if cleaned_text:
        # 3. Detected index exists in model_response, use regex to extract index+comment content
        comments = re.findall(r'(\d+\.\s*.*?)(?=\d+\.|$)', cleaned_text, re.DOTALL)
        return comments
    return []

---example3---
#model_response#: Below is a comment: 1. The environment in the video looks very comfortable, elderly people will definitely like such a place.

#instruction#: Please remove non-comment information or remarks, then output comments from model_response in python list format, split output.

**Important: Only output function: extract_info_list(model_response), only output code! Do not output any other content!**

#extract function#:
def extract_info_list(model_response):
    # 1. No footer remark detected; detected header remark [Below is a comment:], use "^.*comment:" to remove header remark. Don't extract complete header extra remark to ensure regex works properly. Only use few characters after header remark to ensure regex works properly.
    cleaned_text = re.sub(r'^.*comment:', '', model_response, flags=re.DOTALL)
    if cleaned_text:
        # 3. Detected index exists in model_response, use regex to extract index+comment content
        comments = re.findall(r'(\d+\.\s*.*?)(?=\d+\.|$)', cleaned_text, re.DOTALL)
        return comments
    return []

---example4---
#model_response#: beef roasted duck, pork braised carp, chicken fried shrimp, mutton stewed tofu, fish steamed pumpkin

#instruction#: Please remove all non-name information, then split and output all model-generated names

**Important: Only output function: extract_info_list(model_response), only output code! Do not output any other content!**

#extract function#:
def extract_info_list(model_response):
    # No header, footer or middle extra remarks detected, directly use original text
    cleaned_text = model_response
    if cleaned_text:
        # Use comma to split all dish names
        dishes = re.findall(r'[^(*@{\begin{CJK*}{UTF8}{gbsn}、\end{CJK*}}@*)]+', cleaned_text)
        # Remove possible whitespace characters
        dishes = [dish.strip() for dish in dishes]
        return dishes
    return []

---your turn---
#model_response#: {model_response}

#instruction#: {instruction}

**Important: Only output function: extract_info_list(model_response), only output code! Do not output any other content!**

#extract function#:
\end{lstlisting}

\subsection{Single Element General Extract Prompt}
\begin{lstlisting}[escapeinside={(*@}{@*)}]
You are an information extraction expert. You will be provided with an [Input Instruction] and corresponding [Model Response]. Additionally, you will be given an [Extraction Target] designed to evaluate whether the [Model Response] meets certain requirements of the [Input Instruction]. Now, you need to extract the evaluation target from the [Model Response] that corresponds to the [Extraction Target].

**Note: You should completely copy continuous original text segments from the [Model Response], strictly prohibiting any modification, addition, deletion, or splicing, only copying is allowed. Even if it doesn't meet the [Extraction Target] requirements. You don't need to care about what the specific requirements of the [Extraction Target] are, nor do you need to evaluate whether the [Model Response] meets the [Extraction Target] requirements.**

Your purpose is to remove those extra generated auxiliary reading information, such as "Here is the information I provide:", "Great question!"

It is known that there is only one evaluation object in the [Model Response]. Please format it in Python list format and output it.

If the evaluation object is an entire paragraph, output ALL. Try to output ALL to reduce the number of output tokens.

**Please only output a valid Python list evaluation object or ALL, do not output any other notes or any other content!**

---example1---

[Input Instruction]

Market operations qualitative description of the overall situation: clearly explain what key performance results have been achieved, with both quantitative analysis and qualitative summary, avoiding falling into data piling without explaining the meaning behind the data, 200-word summary

[Model Response]

In this quarter, the market operations team successfully drove user growth rate to 18% and improved brand awareness and market share through innovative marketing activities and precise market positioning strategies. Quantitatively, through data analysis, we ensured optimization of marketing input and return rates, while improvements in user activity and engagement also directly reflected the effectiveness of market operations. Qualitatively, our market operations not only enhanced users' brand identification but also improved user engagement through social media and online interaction, strengthening the interaction and emotional connection between brand and consumers. These achievements are not just numerical growth, but represent solid steps we have taken in understanding and meeting consumer needs and building long-term brand value.

[Extraction Target]

Please extract the part that qualitatively describes the overall situation from the model response in Python list format.

**Please only output a valid Python list evaluation object or ALL, do not output any other notes or any other content!**

[Evaluation Object]

ALL

---example2---

[Input Instruction]

The college entrance exam essay topic is: Extinguishing others' lights will not make yourself brighter; blocking others' paths will not make you go further. "One flower blooming alone is not spring, a hundred flowers blooming together fills the garden with spring." If there were only one kind of flower in the world, no matter how beautiful this flower is, it would still be monotonous. The above two materials are from General Secretary Xi Jinping's speeches, using vivid and figurative language to express universal truths. Please write an article based on this, reflecting your understanding and thinking. Requirements: Choose the right angle, determine the theme, clarify the style, draft your own title; do not copy templates, do not plagiarize; do not reveal personal information; no less than 800 words. Below is an essay written by a student based on the topic: All Flowers Fade but One Blooms Alone - The Power of Common Development [Essay content follows...] Evaluate the pros and cons, how many points would it get out of 60? And try to write a perfect college entrance exam essay.

[Model Response]

Evaluation of student essay: Advantages: 1. Clear theme: The article closely follows the material theme, emphasizing the importance of cooperation and sharing, and develops arguments around this core. 2. Clear structure: The article transitions from introduction to body to conclusion, with clear levels and logic. 3. Fluent language: The article uses some rhetorical devices such as metaphor and parallelism, making the article literary. 4. Wide scope: The author elaborated from three levels: national, commercial, and personal, making the content rich and multi-angled. Disadvantages: 1. Insufficient innovation: Although the article has a complete structure, it lacks unique insights and deep thinking, with content being somewhat mediocre. 2. Insufficient specific examples: Although it mentions national cooperation, enterprise cooperation and personal sharing, it doesn't provide specific examples, making the argumentation less powerful. 3. Weak emotional expression: The article is more rational analysis but lacks emotional rendering and deep emotional expression of the theme. Comprehensive score: Based on the above analysis, this essay has complete content and clear structure, but is slightly insufficient in innovation and argumentation strength. If the full score is 60 points, I would give this article about 45 points. It is a well-structured article but needs more unique insights and specific evidence to improve the score. Attempt to write a perfect college entrance exam essay: [Title] Brilliant Light, Shining Because of You [Body] In a vast sea of flowers, each flower blooms in its unique posture, together composing the movement of spring. As General Secretary Xi Jinping said, one person's brilliance may illuminate a corner, but only when stars surround the moon can the entire night sky be illuminated. In the long river of human history, those behaviors of trying to monopolize and admire oneself in isolation are ultimately ephemeral. The wheels of history roll forward, driven by those who understand cooperation and sharing. They understand that extinguishing others' lights not only fails to illuminate their own path but makes the world darker. In international relations, we witness the power of cooperation. When we look back at the Silk Road, it was not only an exchange of goods but also a fusion of cultures and mutual learning of wisdom.

[Extraction Target]

Please extract the perfect essay example part from the model response in Python list format.

**Please only output a valid Python list evaluation object or ALL, do not output any other notes or any other content!**

[Evaluation Object]

["[Title] Brilliant Light, Shining Because of You [Body] In a vast sea of flowers, each flower blooms in its unique posture, together composing the movement of spring. As General Secretary Xi Jinping said, one person's brilliance may illuminate a corner, but only when stars surround the moon can the entire night sky be illuminated. In the long river of human history, those behaviors of trying to monopolize and admire oneself in isolation are ultimately ephemeral. The wheels of history roll forward, driven by those who understand cooperation and sharing. They understand that extinguishing others' lights not only fails to illuminate their own path but makes the world darker. In international relations, we witness the power of cooperation. When we look back at the Silk Road, it was not only an exchange of goods but also a fusion of cultures and mutual learning of wisdom."]

---example3---

[Input Instruction]

The problem of cleaning system junk has always troubled ordinary users. I want to write a research report on this issue, mainly to understand in what scenarios ordinary users encounter this problem; what simple cleaning methods are for them; and whether it's possible to design a fool-proof software to benefit the public. Help me write such a research report, within 3000 words.

[Model Response]

Due to length constraints, the following is a simplified version of the research report outline, which you can further expand as needed: --- # Research Report: System Junk Problems Faced by Ordinary Users and Solutions ## Abstract This report investigates system junk problems encountered by ordinary users in daily computer and mobile device usage. It aims to understand scenarios where users encounter system junk problems, explore simple and effective cleaning methods, and evaluate the feasibility of developing a fool-proof cleaning software. ## Introduction System junk problems refer to temporary files, caches, useless programs and file fragments generated during computer and mobile device usage. These junks occupy storage space, affect device performance, and may even lead to privacy leaks. Ordinary users often lack professional knowledge to effectively manage these system junks. ## Research Methods - Online questionnaire: Distribute questionnaires to ordinary users to collect data. - In-depth interviews: Select some users for in-depth interviews to understand specific usage scenarios. - Market analysis: Research the market situation of existing cleaning tools. ## Scenarios Where Users Encounter System Junk - Software installation and uninstallation: Temporary files generated during program installation, configuration information left after uninstallation. - Web browsing: Browser cache and cookies, etc. - System updates: Old system files not cleaned properly. - Daily use: Temporary files generated during document editing and application usage. ## User Requirements for Cleaning Methods Users generally hope for a simple and intuitive cleaning method, such as: - One-click cleaning function that automatically scans and cleans junk. - Regular cleaning reminders to help users develop good habits. --- Please note that this is only a simplified version of the report outline. The actual research report needs to be written based on research data and analysis-derived specific content, including detailed methodology, data analysis, user feedback, and software design solutions, etc., to meet the 3000-word requirement.

[Extraction Target]

Please extract the entire report from the model response in Python list format.

**Please only output a valid Python list evaluation object or ALL, do not output any other notes or any other content!**

[Evaluation Object]

["Research Report: System Junk Problems Faced by Ordinary Users and Solutions ## Abstract This report investigates system junk problems encountered by ordinary users in daily computer and mobile device usage. It aims to understand scenarios where users encounter system junk problems, explore simple and effective cleaning methods, and evaluate the feasibility of developing a fool-proof cleaning software. ## Introduction System junk problems refer to temporary files, caches, useless programs and file fragments generated during computer and mobile device usage. These junks occupy storage space, affect device performance, and may even lead to privacy leaks. Ordinary users often lack professional knowledge to effectively manage these system junks. ## Research Methods - Online questionnaire: Distribute questionnaires to ordinary users to collect data. - In-depth interviews: Select some users for in-depth interviews to understand specific usage scenarios. - Market analysis: Research the market situation of existing cleaning tools. ## Scenarios Where Users Encounter System Junk - Software installation and uninstallation: Temporary files generated during program installation, configuration information left after uninstallation. - Web browsing: Browser cache and cookies, etc. - System updates: Old system files not cleaned properly. - Daily use: Temporary files generated during document editing and application usage. ## User Requirements for Cleaning Methods Users generally hope for a simple and intuitive cleaning method, such as: - One-click cleaning function that automatically scans and cleans junk. - Regular cleaning reminders to help users develop good habits."]

---your turn---

[Input Instruction]

{input_instruction}

[Model Response]

{model_response}

[Extraction Target]

{extraction_prompt}

**Please only output a valid Python list evaluation object or ALL, do not output any other notes or any other content!**

[Evaluation Object]
\end{lstlisting}

\subsection{Multiple Element General Extract Prompt}
\begin{lstlisting}[escapeinside={(*@}{@*)}]
You are an information extraction expert. You will be provided with an [Input Instruction] and corresponding [Model Response]. Additionally, you will be given an [Extraction Target]. The ultimate purpose of this task is to evaluate whether the [Model Response] meets certain requirements of the [Input Instruction]. However, for now, you don't need to complete the final evaluation - you only need to extract the evaluation objects from the [Model Response] that correspond to the [Extraction Target].

**Note 1: You should completely copy continuous original text segments from the [Model Response], strictly prohibiting any modifications, additions, deletions, or concatenations - only copying is allowed. Even if it doesn't meet the [Extraction Target] requirements. You don't need to care about what the specific requirements of the [Extraction Target] are, nor do you need to evaluate whether the [Model Response] meets the [Extraction Target] requirements.**

**Note 2: You only need to extract content, not any formal additional information, unless there are special requirements in the [Extraction Target]: Common additional information includes: [1. Hello], then only extract "Hello", not "1."; [Answer One: Hello], then only extract "Hello", not "Answer One:"; [The wind will break the waves,], then only extract "The wind will break the waves", not ","; [A. Hello], then only extract "Hello", not "A."**

It is known that there are multiple evaluation objects in the [Model Response]. Please use the following python list format to separate and output them.

["Object1", "Object2", "Object3"]

**Please only output a valid python list of evaluation objects, do not output any other remarks or any other content!**

---example1---

[Input Instruction]
When you browse short videos about elderly care services, from a consumer perspective, output 50 colloquial comments on elderly care service videos. The comments should be mixed in length with half each of long and short comments. Please output long and short comments separately. Short comments should be 1-6 words, long comments should be 10-30 words. Each comment cannot use the same vocabulary and style. Don't ask repeated similar questions; cannot contain words like "this", "you guys", "independent", "gift"; question-type comments should account for 50% of total comments.

[Model Response]
### Short Comments (1-6 words) 1. Really good! 2. Service is so caring. 3. Environment looks elegant. 4. Looks very warm. 5. Seems quite professional. 6. Suitable for elderly living. 7. Interesting activities! 8. Nursing staff are enthusiastic. 9. Feel reassured inside. 10. Facilities are complete. ### Long Comments (10-30 words) 11. The environment in the video looks very comfortable, elderly people would definitely like such a place. 12. I think this elderly care service pays special attention to details, making people feel very secure. 13. Activities are well arranged, can keep elderly people active and social. 14. The nursing team seems very experienced, which is crucial for elderly health. 15. Seeing so many happy smiles shows the atmosphere here is very friendly! 16. For elderly people who need care, such service is indeed a good choice.

[Extraction Target]
Please extract all long comments from the model_response in python list format.

**Please only output a valid python list of evaluation objects, do not output any other explanations, remarks or any other content!**

[Evaluation Objects]
["The environment in the video looks very comfortable, elderly people would definitely like such a place.", "I think this elderly care service pays special attention to details, making people feel very secure.", "Activities are well arranged, can keep elderly people active and social.", "The nursing team seems very experienced, which is crucial for elderly health.", "Seeing so many happy smiles shows the atmosphere here is very friendly!", "For elderly people who need care, such service is indeed a good choice."]

---example2---

[Input Instruction]
Now suppose you are applying for the part-time MPA program at Beihang University and are currently in the interview stage. Your personal information: undergraduate graduate from Northwest A&F University, main work experience at KaiShu Stories and FenBi Civil Service Exam, position as product operations. You need to answer these questions in Chinese, providing three different answers for each question. ##1. You work so busy, how do you balance work and studies ##2. How does our school's MPA help your current work ##3. How does our school's MPA help you achieve future career goals ##4. What will you do if you are not admitted this time

[Model Response]
## 1. You work so busy, how do you balance work and studies Answer One: To balance work and studies, I have developed a detailed time management plan and will use spare time for learning. At the same time, I believe practical problems encountered at work can provide vivid cases for my learning, so work and study can complement each other. Answer Two: I believe good self-management ability is key to handling the balance between work and studies. I will prioritize and plan my work tasks to ensure work quality while leaving enough time to focus on learning. I will also actively communicate with supervisors to seek flexible learning arrangements. Answer Three: I plan to use the flexibility advantages of part-time MPA by adjusting work plans, such as completing work tasks in advance and utilizing weekends and holidays, to ensure sufficient time for learning. Additionally, I will try to bring practical cases and problems from work into learning to achieve organic integration of work and study. ## 2. How does our school's MPA help your current work Answer One: Your school's MPA program will provide me with professional knowledge in public administration, which is very beneficial for my current product operations work. Especially in understanding policy backgrounds, improving user service quality, and optimizing product strategies, MPA learning will directly enhance my professional capabilities at work. Answer Two:

[Extraction Target]
Please extract all answers to "#1. You work so busy, how do you balance work and studies#" from the model_response in python list format.

**Please only output a valid python list of evaluation objects, do not output any other explanations, remarks or any other content!**

[Evaluation Objects]
["To balance work and studies, I have developed a detailed time management plan and will use spare time for learning. At the same time, I believe practical problems encountered at work can provide vivid cases for my learning, so work and study can complement each other.", "I believe good self-management ability is key to handling the balance between work and studies. I will prioritize and plan my work tasks to ensure work quality while leaving enough time to focus on learning. I will also actively communicate with supervisors to seek flexible learning arrangements.", "I plan to use the flexibility advantages of part-time MPA by adjusting work plans, such as completing work tasks in advance and utilizing weekends and holidays, to ensure sufficient time for learning. Additionally, I will try to bring practical cases and problems from work into learning to achieve organic integration of work and study."]

---your turn---

[Input Instruction]
{input_instruction}

[Model Response]
{model_response}

[Extraction Target]
{extraction_prompt}

**Please only output a valid python list of evaluation objects, do not output any other remarks or any other content!**

[Evaluation Objects]
\end{lstlisting}

\subsection{Evaluate Prompt}
\begin{lstlisting}[escapeinside={(*@}{@*)}]
EVALUATION_PROMPT = """
Please act as an objective and fair judge, analyze the content of the [Model Response], and choose "Yes" or "No" to answer whether the subsequent [Sub-question] holds true.

You will be given: [Original Question], [Model Response], [Sub-question], where the [Original Question] is only for reference and not as a judgment criterion. Please note that you only need to consider the [Sub-question].

Please strictly follow this format for output:
Analysis: CORRECT/WRONG xxx
Judgment: Yes/No

Please strictly follow the rule: only consider the [Sub-question].
Please strictly follow the rule: output in English.

---example1---

[Original Question]:
Generate 100 user colloquial comments in the comment section of moving service short videos, including praise, confusion, questions, etc.; comments cannot contain the word "including"; 2-12 words, each comment format must be unique

[Model Response]:
1. Service is super great! 2. Moving speed is fast, high efficiency. 3. Fair price, good value. 4. Careful packaging, items safe. 5. Door-to-door service, convenient and fast. 6. One-stop solution for moving problems. 7. Furniture arranged neatly, satisfied. 8. Easy moving, strong workers. 9. Quality guaranteed, trustworthy. 10. Wide service coverage, good. 11. Will recommend to friends. 12. Staff professional, reliable! 13. Rich service content, meets needs. 14. How to book moving? 15. How to arrange service time? 16. How about door-to-door estimation? 17. Is after-sales service good?

[Sub-question]:
Are the comments each in a unique format?

[Your Analysis and Judgment]:
Note, please strictly follow this format for output:
Analysis: CORRECT/WRONG xxx
Judgment: Yes/No

Please strictly follow the rule: only consider the [Sub-question].
Please strictly follow the rule: output in English.

Analysis: WRONG There are format repetitions in the 100 comments provided in the model response. For example, comments 14 to 100 all adopt similar formats like "Moving service, do you have XX moving service?", only differing in item names. Therefore, these comment formats are not unique.
Judgment: No

---example2---

[Original Question]
Please create 20 multiple choice questions and 15 multiple selection questions based on the following content. The questions need to meet comprehensiveness and mass appeal. Pay attention to format as I need to copy to a table, and include correct answers. Content: Online delivery worker professional code...

[Model Response]
### Single Choice Questions 1. What should online delivery workers comply with? A. National laws and regulations and company rules B. Personal preferences C. Customer demands D. Work habits #Answer# A 2. How should online delivery workers treat their profession? A. Respect B. Despise C. Ignore D. Look down on #Answer# A...

[Sub-question]:
Did the model generate questions?

[Your Analysis and Judgment]:
Note, please strictly follow this format for output:
Analysis: CORRECT/WRONG xxx
Judgment: Yes/No

Please strictly follow the rule: only consider the [Sub-question].
Please strictly follow the rule: output in English.

Analysis: CORRECT The original question required generating 20 single choice and 15 multiple choice questions, which the model answer clearly did not fulfill completely. However, since we only need to consider the [Sub-question], the model response does satisfy the requirement.
Judgment: Yes

---your turn---

[Original Question]:
{input}

[Model Response]
{output}

[Sub-question]:
{question}

[Your Analysis and Judgment]:
Note, please strictly follow this format for output:
Analysis: CORRECT/WRONG xxx
Judgment: Yes/No

Please strictly follow the rule: only consider the [Sub-question].
Please strictly follow the rule: output in English.

"""
\end{lstlisting}

\section{Sample Capability Report}
\label{sec:sample_capa_report}
\begin{lstlisting}[escapeinside={(*@}{@*)}]
{
    "round": 1,
    "meeseeks_score": 0.64526276543322,
    "utility_score": 0.4049586776859504,
    "capability_stats": {
        "Intent Recognition": {
            "percentage": 0.0,
            "correct": 0,
            "wrong": 10,
            "total": 10,
            "children": {
                "Follow instruction under distraction": {
                    "percentage": 0.0,
                    "correct": 0,
                    "wrong": 10,
                    "total": 10,
                    "children": {}
                }
            }
        },
        "Granular Content Validation": {
            "percentage": 0.71640625,
            "correct": 1834,
            "wrong": 726,
            "total": 2560,
            "children": {
                "Theme requirement": {
                    "percentage": 0.9793103448275862,
                    "correct": 994,
                    "wrong": 21,
                    "total": 1015,
                    "children": {}
                },
                "Stylistic requirement": {
                    "percentage": 0.7575757575757576,
                    "correct": 25,
                    "wrong": 8,
                    "total": 33,
                    "children": {
                        "Generate in certain style": {
                            "percentage": 1.0,
                            "correct": 21,
                            "wrong": 0,
                            "total": 21,
                            "children": {}
                        },
                        "Generate names/titles": {
                            "percentage": 0.3333333333333333,
                            "correct": 4,
                            "wrong": 8,
                            "total": 12,
                            "children": {}
                        }
                    }
                },
                "Language requirement": {
                    "percentage": 0.0,
                    "correct": 0,
                    "wrong": 8,
                    "total": 8,
                    "children": {
                        "Generate Chinese-English-mixed article": {
                            "percentage": 0.0,
                            "correct": 0,
                            "wrong": 8,
                            "total": 8,
                            "children": {}
                        }
                    }
                },
                "Granular format requirement": {
                    "percentage": 0.9590017825311943,
                    "correct": 538,
                    "wrong": 23,
                    "total": 561,
                    "children": {
                        "Generate in other format": {
                            "percentage": 0.9549902152641878,
                            "correct": 488,
                            "wrong": 23,
                            "total": 511,
                            "children": {}
                        },
                        "Generate result in date-format": {
                            "percentage": 1.0,
                            "correct": 50,
                            "wrong": 0,
                            "total": 50,
                            "children": {}
                        }
                    }
                },
                "Word count requirement": {
                    "percentage": 0.311336717428088,
                    "correct": 184,
                    "wrong": 407,
                    "total": 591,
                    "children": {
                        "Generate at accurate word number": {
                            "percentage": 0.0,
                            "correct": 0,
                            "wrong": 12,
                            "total": 12,
                            "children": {}
                        },
                        "Generate in rough/range word number": {
                            "percentage": 0.44357976653696496,
                            "correct": 114,
                            "wrong": 143,
                            "total": 257,
                            "children": {}
                        },
                        "Generate in X times word number of reference text": {
                            "percentage": 0.05660377358490566,
                            "correct": 6,
                            "wrong": 100,
                            "total": 106,
                            "children": {}
                        },
                        "Generate in 0~10 words": {
                            "percentage": 0.5517241379310345,
                            "correct": 16,
                            "wrong": 13,
                            "total": 29,
                            "children": {}
                        },
                        "Generate in 10~50 words": {
                            "percentage": 0.6857142857142857,
                            "correct": 24,
                            "wrong": 11,
                            "total": 35,
                            "children": {}
                        },
                        "Generate in 50~200 words": {
                            "percentage": 0.47368421052631576,
                            "correct": 18,
                            "wrong": 20,
                            "total": 38,
                            "children": {}
                        },
                        "Generate in above 200 words": {
                            "percentage": 0.05263157894736842,
                            "correct": 6,
                            "wrong": 108,
                            "total": 114,
                            "children": {}
                        }
                    }
                },
                "Other granular requirements": {
                    "percentage": 0.26420454545454547,
                    "correct": 93,
                    "wrong": 259,
                    "total": 352,
                    "children": {
                        "Generate rhyming content": {
                            "percentage": 0.25,
                            "correct": 4,
                            "wrong": 12,
                            "total": 16,
                            "children": {}
                        },
                        "Generate with certain keywords": {
                            "percentage": 0.0,
                            "correct": 0,
                            "wrong": 42,
                            "total": 42,
                            "children": {}
                        },
                        "Generate repeat/non-repeat content": {
                            "percentage": 0.6056338028169014,
                            "correct": 43,
                            "wrong": 28,
                            "total": 71,
                            "children": {}
                        },
                        "Generate with certain rhetoric": {
                            "percentage": 1.0,
                            "correct": 15,
                            "wrong": 0,
                            "total": 15,
                            "children": {}
                        },
                        "Generate with certain number of word X": {
                            "percentage": 0.14903846153846154,
                            "correct": 31,
                            "wrong": 177,
                            "total": 208,
                            "children": {}
                        }
                    }
                }
            }
        },
        "Output Structure Validation": {
            "percentage": 0.9072,
            "correct": 1134,
            "wrong": 116,
            "total": 1250,
            "children": {
                "Output format requirement": {
                    "percentage": 0.9463203463203463,
                    "correct": 1093,
                    "wrong": 62,
                    "total": 1155,
                    "children": {
                        "JSON format": {
                            "percentage": 0.9463203463203463,
                            "correct": 1093,
                            "wrong": 62,
                            "total": 1155,
                            "children": {}
                        }
                    }
                },
                "Element number requirement": {
                    "percentage": 0.38666666666666666,
                    "correct": 29,
                    "wrong": 46,
                    "total": 75,
                    "children": {}
                },
                "Output logic requirement": {
                    "percentage": 0.6,
                    "correct": 12,
                    "wrong": 8,
                    "total": 20,
                    "children": {
                        "Generate by certain steps": {
                            "percentage": 0.6,
                            "correct": 12,
                            "wrong": 8,
                            "total": 20,
                            "children": {}
                        }
                    }
                }
            }
        }
    },
    "total_items": 363
}
\end{lstlisting}

\end{document}